%% file: main.tex
\pdfoutput=1

\documentclass[10pt,twocolumn,letterpaper]{article}

\usepackage[pagenumbers]{wacv}

\input{preamble}
\definecolor{wacvblue}{rgb}{0.21,0.49,0.74}
\usepackage[pagebackref,breaklinks,colorlinks,allcolors=wacvblue]{hyperref}
\usepackage{orcidlink}

\usepackage{pifont}
\newcommand{\cmark}{\ding{51}}%
\newcommand{\xmark}{\ding{55}}%

\usepackage{algorithm}
\usepackage{algpseudocode}

\usepackage{adjustbox}

\def\wacvPaperID{1431} %
\def\confName{WACV}
\def\confYear{2027}

\title{Unlocking Cross-Attention in Label-Scarce Regimes: A Weakly Supervised Approach for Video Semantic Segmentation}

\author{
Andrea Marelli$^{1}$\,\orcidlink{0009-0009-4418-110X} \qquad Alessandro Ardenghi$^{2}$ \qquad Giacomo Boracchi$^{1}$\,\orcidlink{0000-0002-1650-3054} \qquad Luca Magri$^{1}$\,\orcidlink{0000-0002-0598-8279}\\[4pt]
$^{1}$DEIB, Politecnico di Milano, Italy \qquad $^{2}$Bocconi University, Italy\\
{\tt\small andrea.marelli@polimi.it} \qquad {\tt\small alessandro.ardenghi@studbocconi.it}
}

\begin{document}
\maketitle
\input{sec/all}
{
    \small
    \bibliographystyle{ieeenat_fullname}
    \bibliography{main}
}

\input{sec/X_suppl}

\end{document}

%% file: sec/all.tex
\begin{abstract}
Weakly Supervised Video Semantic Segmentation (WS-VSS) aims to learn temporally consistent semantic object masks in videos using only video-level labels. While fully supervised models successfully learn temporal relationships by means of cross-attention, training these temporal weights under weak supervision is unstable and even pushes the model below its single-frame baseline. Therefore, existing WS-VSS methods exploit temporal consistency by relying on external priors, such as optical flow. Unfortunately these cues are non-semantic, computationally expensive and noisy, degrading the supervisory signal. Instead of supervising attention weights directly or using external priors, we propose MemCAM, a WS-VSS transformer framework that learns temporal cross-attention directly through feature-space self-supervision.  MemCAM introduces a CAM Propagation Algorithm that synthesizes a supervisory signal by propagating past CAMs according to patch-level feature similarities in the latent space.
Specifically, we constrain their induced CAMs through a consistency objective between predicted and propagated maps, stabilizing temporal attention learning.
Experiments on the YouTube-Objects v2.2 and WS$^2$ datasets demonstrate that MemCAM consistently outperforms the temporal localization and segmentation performance of WS-VSS alternatives, showing that cross-attention can be effectively trained in label-scarce video settings through internal self-supervision. Code is available at {\urlstyle{same}\href{https://github.com/andreamarelli99/MemCAM}{\nolinkurl{github.com/andreamarelli99/MemCAM}}}.
\end{abstract}

\section{Introduction}
\label{sec:intro}

Video semantic segmentation (VSS) is becoming increasingly important for a wide range of computer vision applications. Recent foundation models, such as SAM 2 \cite{ravi2024sam} and SAM 3 \cite{carion2025sam}, have achieved remarkable performance in this domain. Their success relies on cross-attention between frames, which lets transformers learn \textit{temporal} relationships much as ViTs \cite{dosovitskiy2020image} learn spatial ones via self-attention. Despite their outstanding performance on natural domains, these foundation models often struggle to generalize to highly domain-specific scenarios like waste sorting \cite{marelli2025ws2, zhang2024low} or industrial defect detection \cite{bovzivc2021mixed, li2025survey}. In such contexts, the dense pixel-level masks required by fully supervised training are costly to acquire. These settings underscore the need for Weakly Supervised (WS) VSS solutions capable of simultaneously learning object semantics and masks using only video-level class or tag. Modern WS methods rely mainly on transformer-based extensions of Class Activation Maps (CAMs) \cite{zhou2016learning} that refine activations by exploiting the \emph{self-attention} weights learned during training on static images \cite{xu2022multi, xu2024mctformer+, zhu2023weaktr}. Although transitioning from self- to inter-frame cross-attention in VSS is \emph{conceptually} straightforward, our experiments show that naively training these temporal cross-attention weights in a label-scarce regime fails: video-level labels provide no signal to optimize inter-frame attention, so the added temporal weights are at best left untrained and at worst drive the model \emph{below} its single-frame baseline (Sec.~\ref{sec:ablation}). Extending this paradigm to WS-VSS thus remains an open problem.

\begin{figure*}[t]
\centering
    \includegraphics[width=0.8\linewidth]{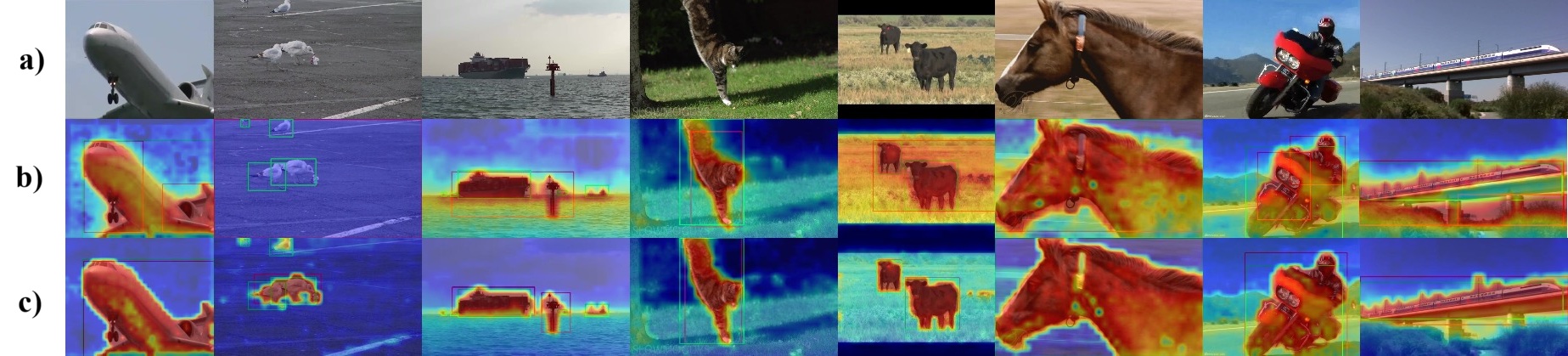}
    \caption{Qualitative comparison of generated CAMs on the \textbf{YouTube-Objects v2.2} dataset. \textbf{(a)} Original video frames. \textbf{(b)} Baseline CAMs generated by the single-frame \textbf{MCTformer+} \cite{xu2024mctformer+} backbone. \textbf{(c)} CAMs generated by our proposed \textbf{MemCAM}. MemCAM produces superior results, demonstrating more precise localization and reduced background noise, achieving more spatially complete and less fragmented results.}
    \label{fig:qualitative_vs_mct_former}
\end{figure*}

To address this challenge, we propose MemCAM, a framework that trains inter-frame cross-attention within this label-scarce regime. Our key insight is that temporal supervision can emerge from feature-space similarity between nearby frames: by matching patch embeddings across frames, we construct a synthetic CAM target that provides stable supervision for temporal attention. In other words, we decouple the temporal context from its supervision, supervising not the cross-attention weights directly but their effect at the CAM level through feature-space propagation. This lets MemCAM learn temporal cross-attention entirely within a WS setting, without dense annotations or external priors such as optical flow or color consistency that existing temporal WS-VSS models rely on \cite{marelli2024temporal, belharbi2023tcam, belharbi2025colo}, and which are non-semantic, fragile to occlusions or lighting variations, noisy, and introduce a significant computational bottleneck. Building on this intuition, our framework introduces three main components:
\begin{itemize}
\item \textbf{Memory Attention Module (MAM)}: A lightweight cross-attention module that retrieves the embeddings of past frames stored in a Memory Bank, allowing the model to refine current features using recent temporal context.
\item \textbf{CAM Propagation Algorithm (CAMPA)}: The core of our method, a feature-matching–based procedure that synthesizes a supervisory CAM for the current frame by propagating CAMs from previous ones, guided by region-level similarities in feature space.
\item \textbf{CAM Consistency Loss}: A loss function that enforces the alignment between the predicted CAM and the synthetic one, serving as the primary self-supervisory signal.
\end{itemize}

Together, these components allow MemCAM to learn temporal coherence from its own feature space, bridging the gap between frame-level weak supervision and temporally consistent video segmentation.
Experiments on YouTube-Objects v2.2 \cite{kalogeiton2016analysing} (temporal localization) and the WS$^2$ \cite{marelli2025ws2} industrial dataset (segmentation), the only two publicly available benchmarks matching our supervision protocol, show that MemCAM sets a new state-of-the-art in WS-VSS. In summary, our main contributions are: (i) We show that inter-frame cross-attention can be effectively trained under video-level supervision, without relying on dense masks or external motion priors, by leveraging learned semantic consistency across frames; (ii) We propose CAMPA, a feature-space CAM propagation strategy that generates synthetic supervision from past frames to stabilize the training of temporal attention in a label-scarce regime; (iii) We introduce a transformer-based WS-VSS architecture that integrates memory-based cross-attention with minimal inference overhead over a standard single-image model. We instantiate it on the family of token-based WS classifiers that natively expose patch tokens and per-class token CAMs (e.g.,~\cite{xu2024mctformer+}), and we delimit and discuss this scope explicitly in Sec.~\ref{sec:supp_scope} of the Supplementary Material.

\section{Problem Formulation}
\label{sec:problem}

WS-VSS aims to learn temporally consistent pixel-level segmentation mask for each frame in a video using only coarse annotations available at video level. Formally, given a training set $\mathbb{T}=\{(v_{i},y_{i})\}_{i=1}^{N}$, where each video $v_{i}=\{I_{t}\in\mathbb{R}^{H\times W\times3}\}_{t=1}^{T_{i}}$ is associated with a single video-level label $y_{i}\in\{1,...,C\}$, our goal is to learn a function $F_{\theta}$ that generates a dense segmentation mask $M_{t}=F_{\theta}(I_{t}) \in[0,1]^{H\times W\times|\mathcal{C}|}$ for each frame $I_{t}$.
Predictions must satisfy two properties:
(i) they should be semantically aligned with the video-level label $y_i$, and
(ii) they should remain temporally consistent across consecutive frames within $v_i$.

\section{Related Work}
\label{sec:related_work}

We review the existing literature along three directions: self-attention for WS image segmentation, cross-attention and temporal reasoning in videos, and WS-VSS methods relying on external motion or appearance cues. To our knowledge, we are the first to learn inter-frame cross-attention for WS-VSS from video-level labels alone, replacing external priors with internal feature-space self-supervision.

\paragraph{Self-Attention for WS Semantic Segmentation (WS-SS) on images.} Rather than relying on CAMs \cite{zhou2016learning} from convolutional classifiers, whose limited receptive field confines them to the most salient object parts \cite{ahn2018learning, wei2017object, ahn2019weakly, lee2021anti, ru2022learning}, ViT's self-attention \cite{dosovitskiy2020image} among image patches allows the network to model long-range dependencies to capture semantic relationships across distant regions, enriching CAMs with semantic and spatial context. Recent transformer-based WS-SS approaches \cite{gao2021ts, xu2022multi, ru2022learning, li2023transcam, zhu2023weaktr} exploit the attention weights learned during training to recover spatial correspondences and localization cues, effectively ``recycling'' self-attention maps as weak saliency priors. Among them, \cite{xu2022multi} introduces a multi-class token design where each learnable token represents a semantic category, allowing the network to attend to multiple objects within a single image and produce CAMs without extra supervision. A parallel line of research instead externalizes the localization signal to a vision--language foundation model: CLIP-based WS-SS methods ~\cite{lin2023clip, zhang2024weclip, yang2025excel, yang2026diclip} align image or patch features with the textual class name to localize categories without dense labels. However, they inherit CLIP's semantics \emph{frozen} at pretraining, and can therefore only localize concepts that fall within CLIP's pretraining distribution. On fine-grained, domain-specific categories that were never seen, this frozen prior is uninformative and costly to adapt or retrain \cite{marelli2025ws2, zhang2024low, bovzivc2021mixed, li2025survey}. Despite their ability to generate semantically richer and more coherent CAMs, none of these methods extend this principle to the video domain, where leveraging cross-attention between frames could further capture temporal relationships and enhance consistency over time.

\paragraph{Cross-Attention and Temporal Reasoning in Videos.}
In VSS, \emph{temporal coherence} is a powerful source of supervision: objects tend to preserve appearance and spatial relationships across consecutive frames. Transformer architectures exploit this through \emph{cross-attention} mechanisms that relate features between the current frame and past ones. Large-scale fully supervised models such as \cite{ravi2024sam, carion2025sam} integrate a memory-based attention module storing key-value pairs of features and masks from previous frames, which the current frame queries through cross-attention to refine its predictions. However, training such temporal relationships requires dense ground-truth (GT) annotations to guide the attention toward correct inter-frame correspondences, making this approach unsuitable for WS contexts. A complementary, unsupervised line of research instead learns inter-frame correspondences directly from raw video, with no dense masks, by leveraging the intrinsic knowledge and temporal robustness of foundation models~\cite{simeoni2025dinov3} to match the attention maps of nearby frames into coherent video segmentations~\cite{caron2021emerging, jabri2020space, ding2024betrayed, delatolas2025diffusion}. These methods are the closest in spirit to ours, as they too recover temporal consistency from feature-space affinities rather than from optical flow. Yet they obtain it by relying on the emergent properties of \emph{frozen} foundation models, which makes them \emph{class-agnostic}: they propagate a generic foreground mask with no notion of category and are hard to adapt to a specific domain. In contrast, we recover temporal consistency while remaining \emph{class-specific}, learning the signal \emph{in-domain} from a fully fine-tuned backbone rather than inheriting it from a frozen prior.

\paragraph{Weakly Supervised Video Semantic Segmentation}
Since training temporal cross-attention without explicit labels is inherently difficult, no existing WS-VSS methods have leveraged transformer-based cross-frame reasoning to improve supervision under weak labeling conditions. Instead, current approaches primarily aim to enforce temporal consistency in convolutional CAM frameworks by propagating activation maps across frames using external signals.
Methods such as ~\cite{belharbi2023tcam, belharbi2025colo, marelli2024temporal} rely on optical flow or color affinity to estimate motion or appearance similarity, which is then used to transfer saliency information between frames. While such priors can enhance temporal smoothness, they are often noisy, computationally expensive, and sensitive to illumination variations or non-rigid motion, limiting their scalability and robustness in real-world video scenarios. MemCAM marries the cross-attention memory mechanism of \cite{ravi2024sam} and \cite{carion2025sam} with the multi-class-token self-attention of \cite{xu2024mctformer+}, recovering temporal consistency while staying WS, class-specific, and domain-trainable.

\section{MemCAM}
\label{sec:memcam}
Our method, MemCAM, is illustrated in Figure \ref{fig:memcam_arch}. It combines the three components introduced in Sec. 1: the \textit{MAM} injects temporal context into the current patch embeddings, \textit{CAMPA} synthesizes the pseudo-GT $\hat{M}^t$, and $\mathcal{L}_M$ aligns the two, since MAM cannot be optimized without CAMPA, as demonstrated in our Ablation (Sec.~\ref{sec:ablation}).
Together, these components transform coarse class supervision into temporally coherent spatial activations. Since we operate in a class-specific setting, inspired by the multi-class token architecture, we designed CAMPA to integrate with an architecture featuring both \emph{class-level} (Fig. \ref{fig:memcam_arch}.a) and \emph{patch-level} (Fig. \ref{fig:memcam_arch}.b) supervision branches. This structural choice exposes both the patch tokens required for feature-space correspondence and the per-class token CAMs to be propagated, thereby preserving spatial structure and semantic alignment. More details of design choices and discussions on alternative architectures are reported in Supp. Sec.~\ref{sec:supp_scope}.

\subsection{Backbone}

\begin{figure*}[t]
    \centering
    \includegraphics[width=0.9\textwidth]{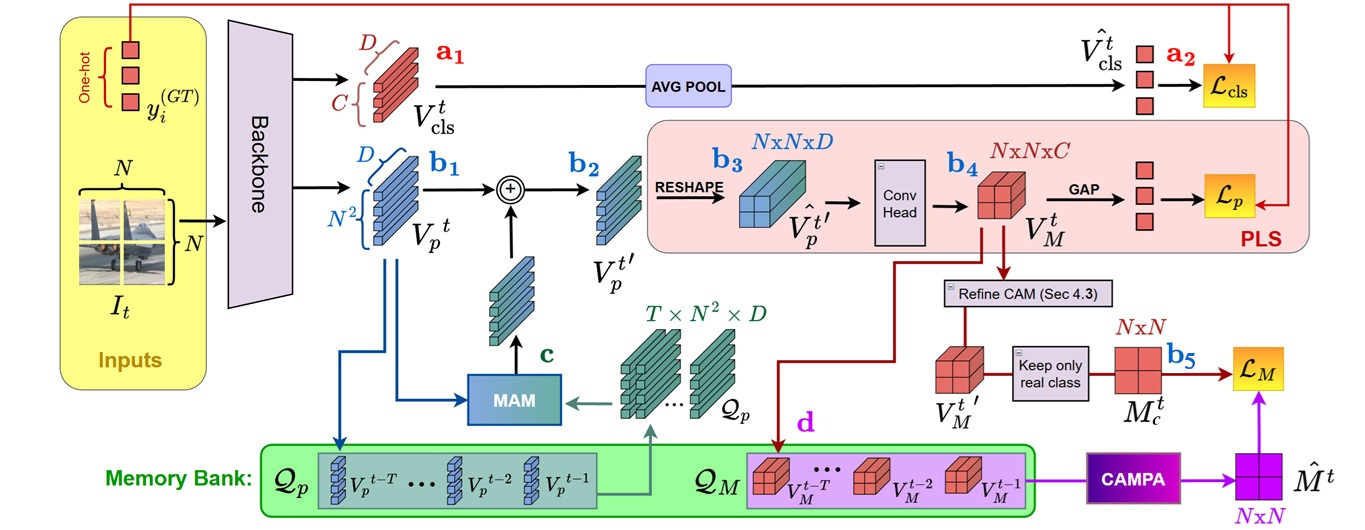}
    \caption{Overview of MemCAM architecture. \textcolor{brown}{Inputs} (left) are encoded by the multi-class transformer backbone into class tokens $V_\text{cls}^t$ (\textcolor{red}{a$_1$}) and patch tokens $V_p^t$ (\textcolor{blue}{b$_1$}). $V_\text{cls}^t$ are pooled and fed to the classification head (\textcolor{red}{a$_2$}) producing $\mathcal{L}_{\text{cls}}$. $V_p^t$ are refined via the MAM, (\textcolor{green}{c}) using context retrieved from the Memory Bank's patch queue $\mathcal{Q}_p$. The memory-refined patch embeddings ${V_p^t}'$ (\textcolor{blue}{b$_2$}) are reshaped (\textcolor{blue}{b$_3$}) and passed through the Patch-Level Supervision branch (\textcolor{red}{PLS}) to produce per-patch activations $V_M^t$ and the patch-level loss $\mathcal{L}_{p}$. Class-aware CAMs ${V_M^t}'$ are obtained after attention-based refinement and stored in the CAM queue $\mathcal{Q}_M$ of the Memory Bank (\textcolor{violet}{d}). \textcolor{violet}{CAMPA} consumes the CAMs stored in $\mathcal{Q}_M$ to produce a propagated pseudo-CAM $\hat{M}^t$, which is used to compute the consistency loss $\mathcal{L}_{\mathrm{M}}$.}
    \label{fig:memcam_arch}
\end{figure*}

The backbone's role is to encode each input frame $I_t$ at time $t$ (divided into $N \times N$ patches) into two complementary representations:
(i) \emph{class tokens} $V_{\text{cls}}^t \in \mathbb{R}^{C \times D}$ (Figure~\ref{fig:memcam_arch}\textbf{a$_1$}), which summarize the semantic content of the frame, and
(ii) \emph{patch tokens} $V_{p}^t \in \mathbb{R}^{N^2 \times D}$ (Figure~\ref{fig:memcam_arch}\textbf{b$_1$}), which capture fine-grained spatial features and are later refined through temporal attention (see Sec.~\ref{sec:mem_attn}–\ref{sec:patch_supervision}).
Here, $D$ denotes the feature dimension and $C$ the number of semantic classes. During training, the class tokens $V_{\text{cls}}^t$ are mean-pooled across the embedding dimension to obtain $\hat{V}_{\text{cls}}^t$, which is then passed through a cross-entropy loss:
\begin{equation}
\mathcal{L}_{\text{cls}} = \text{CrossEntropy}(\hat{V}_{\text{cls}}^t, y_i^{GT}),
\end{equation}
as shown in Figure~\ref{fig:memcam_arch}\textbf{a$_2$}.
This loss enforces semantic alignment between the global video label and the internal feature representations learned by the network (Figure~\ref{fig:memcam_arch}\textbf{a$_2$}).

\subsection{Memory Attention Module}
\label{sec:mem_attn}
MCTformer+ encodes each frame independently, without modeling temporal relationships.
To incorporate information from previous frames and enforce temporal consistency across predicted masks, we introduce the \textbf{Memory Attention Module (MAM)} (Figure~\ref{fig:memcam_arch}\textbf{c}), a lightweight transformer block that refines the embeddings of the current frame using contextual information stored from past frames.
This mechanism relies on a \textbf{Memory Bank} (Figure~\ref{fig:memcam_arch}\textbf{d}) composed of two synchronized FIFO queues of fixed length $T$:
(i) a patch queue $\mathcal{Q}_p$, which stores the patch embeddings $V_p^{t-\tau}$ from the $T$ most recent frames (Figure~\ref{fig:memcam_arch}\textbf{b}); and
(ii) a CAM queue $\mathcal{Q}_M$, which holds the corresponding class activation maps (Figure~\ref{fig:memcam_arch}\textbf{d}) and is further detailed in Section~\ref{sec:cam_loss}. The Memory Bank is \emph{parameter-free}: it stores $T\!\cdot\!N^2\!\cdot\!D$ values for the patch queue and $T\!\cdot\!N^2\!\cdot\!C$ for the CAM queue, i.e.\ under $0.8$\,MB in fp32 at training resolution and about $10$\,MB per scale at the $800\times800$ inference cap. Its cost is linear in $T$ and independent of video length. At each time step, the current patch tokens $V_p^t$  (Figure~\ref{fig:memcam_arch}\textbf{b$_1$}) query the patch queue $\mathcal{Q}_p$  to align the semantic features of frame $t$ with those of its temporal neighbors in the Memory Bank, yielding refined features $V^{t'}_p$ (Figure~\ref{fig:memcam_arch}\textbf{b$_2$}). Formally, the patch queue is defined as
\begin{equation}
    \mathcal{Q}_p = [\, V_p^{t-T}, \ldots, V_p^{t-2}, V_p^{t-1} \,]
    \in \mathbb{R}^{T \times N^2 \times D},
    \label{eq:patch_queue}
\end{equation}
representing a memory of the most recent $T$ frames. The MAM builds upon the attention mechanism~\cite{vaswani2017attention} and operates in three stages:
\emph{(a) intra-frame self-attention} and
\emph{(b) inter-frame cross-attention}
\emph{(c) feed-forward refinement}.

\paragraph{a) Intra-frame Self-Attention}
In this stage, the model computes self-attention among all the patch tokens extracted from the same frame. Learnable linear projections produce the following triplet
\[
\textbf{q}^{\text{self}} = V_p^t W_\textbf{q}^{\text{self}}, \quad \textbf{k}^{\text{self}} = V_p^t W_\textbf{k}^{\text{self}} , \quad \textbf{v}^{\text{self}} = V_p^t W_\textbf{v}^{\text{self}},
\]
where $W_\textbf{q}^{\text{self}}, W_\textbf{k}^{\text{self}}, W_\textbf{v}^{\text{self}} \in \mathbb{R}^{D \times D}$ are learnable weight matrices. The self-attention output is computed as $ A_t^{\text{self}}= \text{Attn}(\textbf{q}^{\text{self}},\textbf{k}^{\text{self}},\textbf{v}^{\text{self}}) \in \mathbb{R}^{N^2 \times D}$ from \cite{vaswani2017attention} (Supp.\ref{sec:mem_attn_details}), allowing each patch to aggregate information from all others within the same frame.
A residual connection is then applied:
\[
{V_p^t}^\text{self} = V_p^t + A_t^{\text{self}}.
\]

\paragraph{b) Inter-frame Cross-Attention}
To incorporate temporal context from past frames, we compute an additional attention operation between the current patch embeddings $V_p^t$ and the embeddings stored in the patch queue~$\mathcal{Q}_p$ (Eq.~\ref{eq:patch_queue}). Each token is associated with the frame it was extracted from by adding absolute sinusoidal temporal positional embeddings~\cite{vaswani2017attention}, thus encoding temporal ordering of frames during attention computation. Let $PE_t \in \mathbb{R}^{N^2 \times D}$ be the encoding for the current frame ($t=0$), and $PE_{\text{mem}}=[PE_\tau]_{\tau=t-T}^{t-1}  \in \mathbb{R}^{T \times N^2 \times D}$ the stacked positional embeddings of the $T$ past frames. The cross-attention $A_t^{\text{cross}}$ between frames is computed as $A_t^{\text{cross}}=\text{Attn}(\textbf{q}^{\text{cross}}, \textbf{k}^{\text{cross}}, \textbf{v}^{\text{cross}})$ where:
\[
\textbf{q}^{\text{cross}} = ({V_p^t}^{\text{self}} + PE_t)W_\textbf{q}^{\text{cross}},
\]
\[
\textbf{k}^{\text{cross}} = (\mathcal{Q}_p + PE_{\text{mem}}) W_\textbf{k}^{\text{cross}} , \qquad
\textbf{v}^{\text{cross}} =\mathcal{Q}_p  W_V^{\text{cross}}.
\]
This mechanism enables each patch in the current frame to query semantically related patches from previous frames, transferring contextual and temporal information across time.
Another residual update follows:
\[
{V_p^t}^\text{cross} = {{V_p^t}^\text{self}} + A_t^{\text{cross}}.
\]

\paragraph{c) Feed-Forward Refinement.}
Finally, ${V_p^t}^\text{cross}$ is processed by a two-layer multi-layer perceptron $MLP(\cdot)$, combined with a final residual connection, producing:
\[
{V_p^t}^{\text{mem}} = MLP({V_p^t}^\text{cross}) + {V_p^t}^\text{cross}.
\]
These embeddings preserve the spatial structure of the original patch tokens while integrating temporal context retrieved through attention to the memory bank.
To obtain the final memory-refined patch embeddings ${V_p^t}'$, we apply a last residual connection directly with the backbone output:
\begin{equation}
\label{eq:mem_ref_pathes}
    {V_p^t}' = V_p^t + {V_p^t}^{\text{mem}} \in \mathbb{R}^{N^2 \times D}.
\end{equation}

\subsection{Patch-level supervision}
\label{sec:patch_supervision}

Class-level supervision says \textit{what} is in the frame but not \textit{where}. We therefore use the MCTformer+ \emph{Patch-Level Supervision branch (PLS)} (Figure \ref{fig:memcam_arch} \textbf{PLS}) to transform the temporally refined patch embeddings from (\ref{eq:mem_ref_pathes}) into spatially distributed activation scores. The PLS encourages the model to produce features that are both class-discriminative and localized. Given the memory-refined patch embeddings ${V_p^t}'$ (Eq. \ref{eq:mem_ref_pathes})  obtained from the MAM, the network predicts a class activation score for each patch. These patch-level class-score predictions are then used to compute an auxiliary classification loss to reinforce the alignment between spatial features and the video-level label $y_i$. Concretely, we first restore the 2D spatial structure of the tokens from (\ref{eq:mem_ref_pathes}):
\[
\widehat{V}_p^{t} = \operatorname{reshape}\!\big({V_p^t}'\big) \in \mathbb{R}^{N \times N \times D},\quad \text{(Figure \ref{fig:memcam_arch}\textbf{b$_3$})}
\]
where $\widehat{V}_p^{t}[i,j,:]$ denotes the $D$-dimensional embedding of the patch at spatial location $(i,j)$. A lightweight convolutional head $f_{\text{conv}}(\cdot)$ maps these embeddings to per-class activation scores at patch resolution:
\begin{equation}
\label{eq:V_M}
V_M^{t} = f_{\text{conv}}\!\big(\widehat{V}_p^{t}\big) \in \mathbb{R}^{N \times N \times C}.\quad \text{(Figure \ref{fig:memcam_arch}\textbf{b$_4$})}
\end{equation}
We denote the per-patch, per-class score at location $(i,j)$ by $V_M^{t}[i,j,c]$.
A spatial GAP is then applied over $V_M^t$ to obtain frame-level class logits, which are compared against the GT video label $y$ using another cross-entropy loss
\begin{equation}
\mathcal{L}_{\text{p}} = \text{CrossEntropy}(GAP(V_M^t), y_i^{GT}),
\end{equation}
This ties spatial embeddings to semantic labels, yielding patch embeddings to be both temporally coherent and spatially informative. The resulting  $V_M^t$  acts as a \emph{patch-level CAM}, to be used for the CAM refinement and propagation.

Finally, to obtain the CAM $M_c^t$ for class $c$ in frame $I_t$, we apply a \textit{class-specific attention refinement} that merges the patch-to-patch self-attention weights $A_t^{\text{self}}$ with the memory-refined patch embeddings $V_M^t$, following the formulation of~\cite{xu2024mctformer+}. The resulting class-specific attention-refined embeddings are denoted as ${V_M^t}'$. By selecting from ${V_M^t}'$ only the channel corresponding to the GT class $c$, we derive the final CAM $M_c^t$ (Fig. \ref{fig:memcam_arch}.b$_5$).

\begin{figure*}[t]
\centering
 \makebox[\textwidth][c]{%
 \includegraphics[width=0.9\textwidth]{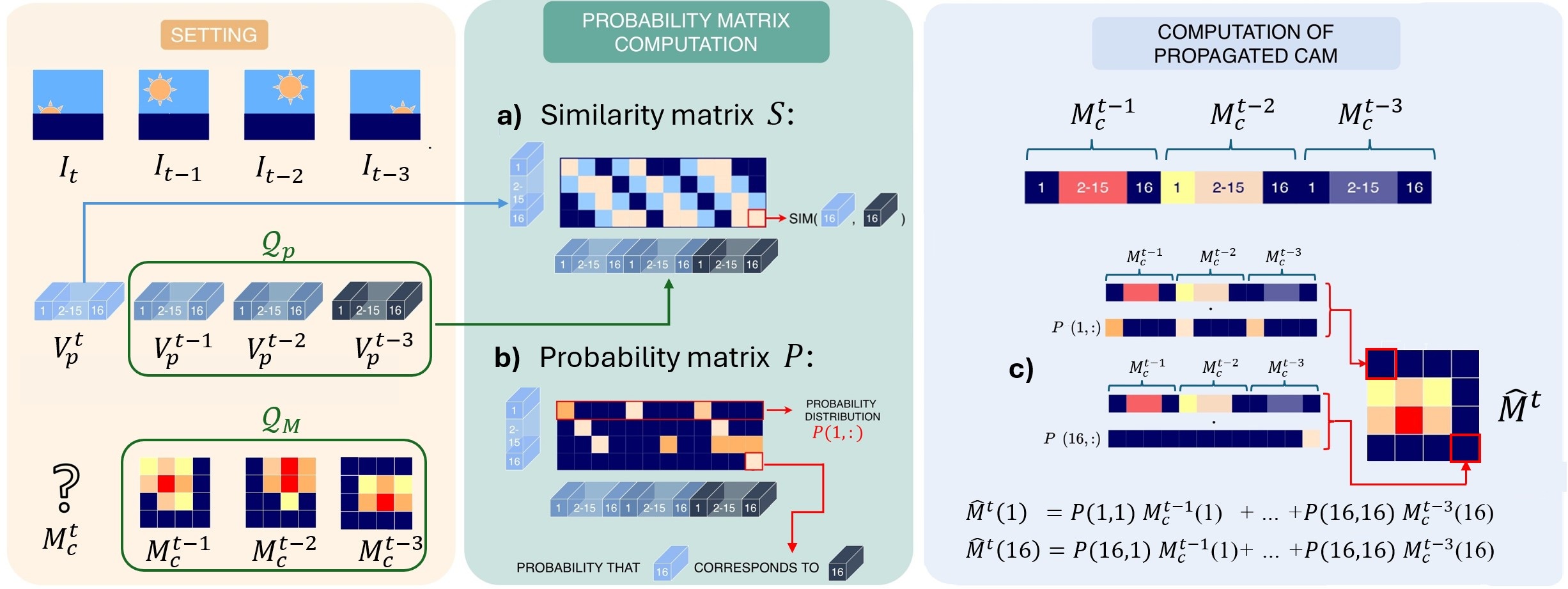}%
 }
\caption{\textbf{CAMPA overview:} The current frame's patch embeddings $V_p^t$ and the memory queues from past frames patch queue $\mathcal{Q}_p$ and CAM queue $\mathcal{Q}_M$ as input (section \textbf{Setting}). \textbf{(a)} A \textbf{Similarity Matrix S} is computed by comparing the current frame's patch embeddings against all embeddings stored in the $\mathcal{Q}_p$ queue. \textbf{(b)} We convert S into a \textbf{Probability Matrix P}, by keeping only the top K entries per row (and so per current patch) and then applying a per-row softmax. In this way, we can see P as a matching likelihood between current and past patches. \textbf{(c)} The final propagated CAM $\hat{M}^t$ is computed as a weighted aggregation of past CAMs (from $\mathcal{Q}_M$), using the probabilities from matrix P to weight the activation values. $\hat{M}^t$ will be used as pseudo-label to supervise the CAM of the current frame $t$. A step-by-step pseudo-code is given in Alg.~\ref{alg:campa} in Section \ref{sec:supp_campa} the Supp. Mat.}
 \label{fig:cam_propagation_algorithm}
\end{figure*}

\subsection{CAM Propagation Algorithm (CAMPA)}
\label{sec:cam_loss}
While the MAM introduces temporal context at the feature level, it does not explicitly enforce consistency between the activation maps of consecutive frames. Although $M_c^t$ represents a valid CAM for frame $t$, our experiments (Sec. ~\ref{sec:experiments}) show that relying solely on the MAM, without any additional supervision, degrades CAM quality compared to not using memory at all. Our \emph{CAM Propagation Algorithm} addresses this by transferring activation patterns from previous frames to the current one, promoting temporal coherence at the mask level through self-supervision. Intuitively, if the backbone produces high-quality embeddings, semantically similar patches across frames lie close in feature space. We exploit this property to transfer class evidence across time, propagating past CAM activations into the current frame according to their inter-frame patch similarities.

As illustrated in Figure~\ref{fig:cam_propagation_algorithm}, CAMPA constructs a synthetic class activation map $\hat{M}_c^t$ for each frame $I_t$ by reweighting and aggregating the CAMs of the $T$ preceding frames according to feature-space similarity. The resulting synthetic map acts as a pseudo-label, guiding the model to maintain coherent object activations over time. A consistency loss between $\hat{M}_c^t$ and the model’s predicted CAM $M_c^t$ (Fig. \ref{fig:memcam_arch}.b$_5$) provides an additional self-supervised signal. As shown in Figure~\ref{fig:memcam_arch}\textbf{d}, once class-aware embeddings ${V_M^t}'$ are computed in (\ref{eq:V_M}), they are stored in the Memory Bank dedicated CAM queue $\mathcal{Q}_{\text{M}} \in \mathbb{R}^{T \times M \times C}$, a FIFO structure parallel to the patch queue $\mathcal{Q}_{p}$. This dual-memory design allows the model to retain both patch embeddings and their corresponding CAMs from the past $T$ frames. Let \(V_p^t\in\mathbb{R}^{N^2\times D}\) be the current-frame patch embeddings and \(\mathcal{Q}_p=[V_p^{t-T},\dots,\) \(V_p^{t-1}]\in\mathbb{R}^{T\times N^2\times D}\) the patch queue from (\ref{eq:patch_queue}).
We compute frame-wise cosine affinities between the current patches and each past frame patch embeddings, applying a temporal decay \(\lambda^\tau\) to emphasize recent frames:
\begin{equation}
\label{eq:comp_sim_submatrixes}
s_{t-\tau} \;=\; \lambda^\tau \,\widehat{V}_p^{\,t}\,(\widehat{V}_p^{\,t-\tau})^\top \in\mathbb{R}^{N^2\times N^2},
\text{  } \tau=1,\dots,T
\end{equation}
where \(\widehat{V}\) denotes $\ell_2$-normalized embeddings along $D$. The per-frame affinity matrices are concatenated into a single similarity matrix $S = \big[\, s_{t-T}, \dots, s_{t-1} \,\big] \in \mathbb{R}^{N^2\times (T N^2)}$  (Fig. \ref{fig:cam_propagation_algorithm}\textbf{a}).
To reduce spurious matches we retain only the Top-\(K\) entries per row (set other entries to \(-\infty\)) and then apply a temperature-scaled, row-wise softmax to obtain a matching probability matrix \(P \in [0,1]^{N^2\times (T N^2)}.\) (Fig. \ref{fig:cam_propagation_algorithm}\textbf{b}). Let \(M_c^{t-\tau}\in\mathbb{R}^{N^2}\) (\(\tau=1,\dots,T\)) denote the past class-specific CAMs, obtained by selecting the GT class channel $c$ from the embeddings ${V_M^{t-\tau}}'$ stored in $\mathcal{Q}_M$. The synthetic CAM for the current frame is then computed as the similarity-weighted aggregation of these past activations:
\begin{equation}
\label{eq:propagated_cam}
\hat{M}^t(i) \;=\; \sum_{\tau=1}^{T}\sum_{j=1}^{N^2} P(i,(\tau -1 )N^2 +j)\,M_c^{\,t-\tau}(j),
\end{equation}
where $P(i, (\tau -1)N^2 + j)$ indicates the probability that patch $i$ of the current frame corresponds to patch $j$ of frame $\tau$ (Fig. \ref{fig:cam_propagation_algorithm}\textbf{b}). The map \(\hat{M}^t\) then supervises the model’s predicted CAM \(M_c^t\) via the consistency loss. A pseudo-code of CAMPA is given in Alg.~\ref{alg:campa} of the Supp., Sec. \ref{sec:supp_campa}

\paragraph{Consistency Loss Computation.}
Considering $M_c^t \in \mathbb{R}^{N^2}$ as described in Section~\ref{sec:patch_supervision}, and $\hat{M}^t$ the corresponding propagated (synthetic) CAM, we define the consistency loss $\mathcal{L}_{\mathrm{M}}$ as the Kullback–Leibler (KL) divergence between $\hat{M}^t$ and $M_c^t$, after $\ell_1$ normalization of both maps so they can be interpreted as probability over spatial locations. Intuitively, CAMPA acts as a \emph{temporal teacher}: the model reuses its own past CAMs to supervise the current frame, so reliable past activations are propagated forward while a locally noisy activation is pulled toward the more reliable consensus of recent frames. While using $\hat{M}^t$  directly as the output matches the student prediction to within $0.7$ CorLoc and $0.4$ mIoU, we use it as a training target instead of a predictor since it would require the CAM queue and the per-frame matching at inference and would leave the first $T$ frames of each video undefined (Sec.~\ref{subsec:supp_campa_as_pred}).

\subsection{Overall Loss}
Let $\lambda_{\text{cls}}, \lambda_{\text{p}}, \lambda_{\text{M}}$ be loss-balancing hyperparameters tuned empirically. During training, the overall loss computed in the forward pass is:
\begin{equation}
\label{eq:final_loss}
\mathcal{L} = \lambda_{\text{cls}}\mathcal{L}_{\text{cls}} + \lambda_{\text{p}}\mathcal{L}_{\text{p}} + \lambda_{\text{M}} \mathcal{L}_{\text{M}}.
\end{equation}
Full derivations of the MAM (Sec.~\ref{sec:mem_attn_details}) and CAMPA (Sec.~\ref{sec:supp_campa}), with the training-stability analysis, the $\ell_1$-vs-minmax normalization, gradient flow through CAM refinement are provided in the Supp. Mat. (Sec.~\ref{sec:training_stability}).

\section{Implementation Details}
 We conducted experiments using both the \textbf{DeiT-Small} (DeiT-S) and \textbf{DeiT-Tiny} (DeiT-T) \cite{touvron2021training} architectures as backbones. The backbone is initialized with ImageNet \cite{deng2009imagenet} weights for fair comparison with SoTA, while we leave the memory attention randomly initialized. Input video frames are resized to $256 \times 256$ pixels and then random-cropped to the standard transformer input resolution of $224 \times 224$. Training is composed of two steps. First, we train a MCTformer+ model keeping only the classification losses active. In the second phase, we initialize MemCAM with the weights of the trained backbone. Additionally, we use the backbone trained in the previous step as a teacher for MemCAM. Indeed, the teacher fills MemCAM's memory with $T$ consecutive CAMs, and then MemCAM uses them as context in the Memory Attention. Furthermore, the $T$ teacher-generated CAMs are used to generate the synthetic CAM through CAMPA. During this second phase, the teacher remains frozen, while the student learns only under the CAM Consistency Loss. Training in both steps are run for 10 epochs with early stopping, AdamW optimizer with learning rate of $5\times10^{-4}$, weight decay $5\times10^{-2}$, and batch size $32$. The \textbf{Memory Bank} stores the patch embeddings and CAMs from the $T=5$ preceding frames, balancing temporal context with storage constraints; performance is flat for $T\in\{2,3,5,7\}$ and degrades only for longer memories, as ablated in Sec.~\ref{subsec:supp_memory_length} of the Supp. Mat. During inference, images are capped at a maximum resolution of $800 \times 800$ pixels while preserving the aspect ratio. Final segmentation masks are computed by postprocessing the CAMs with gamma correction, minmax normalization and hysteresis thresholding for YouTube-Objects v2.2, and simple minmax normalization and thresholding at $\tau=0.75$ for WS$^2$. The complete training and inference protocols, including the sequential dataloaders, the clip-consistent augmentation strategy, the Teacher--Student scheme, and the mask-extraction pipeline, are detailed in Sec.~\ref{sec:supp_training} of the Supp.

\begin{table}[t]
\centering
\resizebox{\columnwidth}{!}{%
\begin{tabular}{cccccc}
\toprule
\textbf{Memory (M)} & \textbf{CAM Loss ($\mathcal{L}_{\text{CAM}}$)} & \textbf{Frozen (F)} & \textbf{Test CorLoc (\%)} & \textbf{Params (M)} & \textbf{Time (ms)} \\
\midrule
\xmark & \xmark & \xmark & 68.6 & 5.7 & 10.9 \\
\cmark & \xmark & \xmark & 67.7 & 6.2 & 11.2 \\
\xmark & \cmark & \xmark & 74.8 & 5.7 & 12.9 \\
\cmark & \cmark & \xmark & \textbf{76.8} & 6.2 & 13.3 \\
\cmark & \cmark & \cmark & 70.7 & 6.2 & 13.3 \\
\bottomrule
\end{tabular}%
}
\caption{Ablation study on the contributions of Memory (M), CAM consistency loss ($\mathcal{L}_{\text{CAM}}$), and backbone freezing (F), with parameter count and per-frame inference time (same measurement protocol as Tab.~\ref{tab:ytov22_params}). Crucially, these results demonstrate that cross-attention alone is not learnable under weak supervision without a synthetic supervisory signal. Over the single-frame baseline the MAM adds $0.3$\,ms and $0.5$M parameters and the CAM loss $2.0$\,ms with none, for a total of $+2.4$\,ms: a $22\%$ inference overhead for $+8.2$ CorLoc. Results from MemCAM$_{\text{tiny}}$ applied on YTOv2.2.}
\label{tab:loss_ablation}
\end{table}

\section{Experiments}
\label{sec:experiments}
In this section, we present ablation studies and quantitative evaluations on two datasets with video-level supervision, comparing MemCAM against state-of-the-art weakly supervised video segmentation and localization approaches.

\subsection{Ablation Studies}
\label{sec:ablation}
We compare two training regimes: freezing the backbone and training only the Memory Attention and head and finetuning the entire network end-to-end. We also compare different Memory Attention depths \(L\in\{1,2,3\}\) and two loss configurations: \emph{Consistency-only} (\(\lambda_{\mathrm{CLS}}{=}0,\ \lambda_{\mathrm{PATCH}}{=}0,\ \lambda_{\mathrm{CAM}}{=}100\)) and \emph{CE-only} ( \( \lambda_{\mathrm{CLS}}{=}1,\ \lambda_{\mathrm{PATCH}}{=}1,\ \lambda_{\mathrm{CAM}}{=}0\)). Table~\ref{tab:loss_ablation} ablates the three ingredients of MemCAM on YTOv2.2 with the Tiny backbone: the Memory Attention (M), the CAM consistency loss ($\mathcal{L}_{\mathrm{CAM}}$), and backbone freezing (F). Crucially, training the network with the MAM \emph{without} any synthetic CAM supervision (\emph{CE-only}, Table~\ref{tab:loss_ablation}, row 2) is in fact worse than the static baseline ($67.7$ vs.\ $68.6$ CorLoc, row 1). This provides direct empirical evidence for our core claim: \textbf{cross-attention alone is not learnable under weak supervision without a synthetic supervisory signal}. Relying solely on class-level labels is insufficient to optimize the inter-frame cross-attention; without proper supervision, this leads to negative transfer and degraded spatial activations. Conversely, training with only the CAM consistency loss, without MAM, significantly improves performance, demonstrating that our self-supervision algorithm and CAM loss are highly effective. Importantly, adding MAM, CAMPA and CAM loss together yields further gains, indicating a synergy between Memory Attention and temporal consistency loss, confirming that MemCAM provides a successful strategy to achieving high-quality segmentation and training frame-wise cross-attention using only self-supervision. Further ablations, isolating the individual CAMPA design choices (Top-$K$ sparsification, the temporal decay $\lambda$, and the softmax temperature $\mu$, the memory-attention depth $L$, and data augmentations) are reported in Sec.~\ref{sec:ablation_supp} of the Supp. Mat. We also show that this gain does not come from a better classifier: classification accuracy \emph{decreases} on both benchmarks while localization improves, consistently with the known decoupling between the two objectives (details in Sec.~\ref{subsec:supp_classification}).

\subsection{Main Experimental Results}
\label{sec:datasets}

We evaluate our approach on the only two publicly available datasets with video-level annotations (more details are reported in Sec.~\ref{sec:supp_failure} of the supplementary material), covering both general and domain-specific contexts:

\begin{itemize}
    \item \textbf{YouTube-Objects v2.2 (YTOv2.2)} \cite{kalogeiton2016analysing}: Although originally designed for WS video object localization (WSVOL), YTOv2.2 is the \textbf{only general-domain dataset} providing video-level labels. We include it to assess the generalization of our method and to compare fairly with WSVOL approaches. Performance is reported using the standard CorLoc metric \cite{belharbi2023tcam}, which quantifies the percentage of correctly localized object regions.
    \item \textbf{WS$^2$} \cite{marelli2025ws2}: A domain-specific industrial dataset collected in a recycling plant and the only existing dataset effectively designed for WS-VSS. It comprises two classes representing scenes \emph{before} and \emph{after} the removal of recyclable materials from a conveyor belt. The task involves segmenting the target objects in the “before” frames, demonstrating the effectiveness of WS-VSS in practical industrial inspection scenarios. For WS$^2$, segmentation performance is measured using mean IoU. An extended analysis of WS$^2$, its intrinsic difficulty, and MemCAM's behaviour on it is provided in Sec.~\ref{sec:supp_ws2} of the Supp. Mat..
\end{itemize}

For both datasets, we divide existing approaches along two main dimensions: (i) methods that operate on static images without leveraging temporal information, and (ii) methods that exploit external cues to enforce temporal consistency.
Among image-based WS approaches, we primarily compare with the baseline of our architecture ~\cite{xu2024mctformer+}, and ~\cite{jo2021puzzle}, which promotes self-supervised CAM refinement by enforcing consistency between multiple CAM generation branches.
All the WS-VSS methods currently available ~\cite{belharbi2023tcam, belharbi2025colo, marelli2024temporal} propagate activation maps across time by leveraging externally computed cues such as optical flow or color affinity.
In contrast, MemCAM learns temporal consistency directly in the latent space, being the first WS framework to combine \emph{self-} and \emph{cross-attention} without relying on external temporal priors. We evaluate two model variants built upon the DeiT backbone~\cite{touvron2021training}:
\textbf{MemCAM\_Tiny}, using the DeiT-T architecture (5.7M param.), and \textbf{MemCAM\_Small}, based on DeiT-S (22M param.).
They share identical training hyperparameters and differ only in backbone capacity.

\begin{table}[t]
    \centering
    \begin{tabular}{lccc}
    \toprule
    \textbf{Method} & \textbf{CorLoc} & \textbf{Params} & \textbf{Time} \\
    \midrule
    \multicolumn{4}{l}{\textit{Static, image-based}}\\
    CAM \cite{zhou2016learning}               & 51.5 & 23.5 & 0.2 \\
    GradCAM \cite{selvaraju2017grad}          & 54.5 & 23.5 & 27.8 \\
    GradCAM++ \cite{chattopadhay2018grad}     & 61.2 & 23.5 & 28.0 \\
    Smooth-GradCAM++ \cite{omeiza2019smooth}     & 59.5 & 23.5 & 136.2 \\
    XGradCAM \cite{fu2020axiom}               & 54.6 & 23.5 & 14.2 \\
    LayerCAM \cite{jiang2021layercam}         & 66.0 & 23.5 & 17.9 \\
    F-CAM \cite{belharbi2022f}                & 67.6 & 23.5 & 18.5 \\
    CLIP-ES \cite{lin2023clip}                & 60.2 & $\sim$100\textsuperscript{*} & 49.9 \\
    MCTformer+-tiny \cite{xu2024mctformer+}               & 68.6 & 5.7  & 10.9 \\
    MCTformer+-small \cite{xu2024mctformer+}              & 71.1 & 21.6 & 11.0 \\
    \midrule
    \multicolumn{4}{l}{\textit{Temporal, external cues}}\\
    TCAM + LayerCAM \cite{belharbi2023tcam}      & 72.2 & 32.5 & 18.5 \\
    CoLo + LayerCAM \cite{belharbi2025colo}      & 77.1 & 32.5 & 18.5 \\
    \midrule
    \multicolumn{4}{l}{\textit{Temporal, self-supervised (ours)}}\\
    MemCAM-tiny                                   & 76.8 & 6.2  & 13.3 \\
    MemCAM-small                                  & \textbf{78.3} & 26.4 & 13.4 \\
    \bottomrule
    \end{tabular}
    \caption{Average CorLoc (\%), parameter count (M) and inference time (ms per frame) for various methods on the \textbf{YTOv2.2} dataset. *Approximation for CLIP-ES (Transformer backbone). Values measured with ResNet50 or comparable backbones.}
    \label{tab:ytov22_params}
\end{table}

\begin{table}[t]
    \centering
    \begin{tabular}{lccc}
    \toprule
    \textbf{Method} & \textbf{mIoU} & \textbf{Params} & \textbf{Time} \\
    \midrule
    \multicolumn{4}{l}{\textit{Static, image-based}}\\
    GradCAM \cite{selvaraju2017grad}      & 19.86 & 23.5 & 27.8 \\
    GradCAM++ \cite{chattopadhay2018grad} & 16.40 & 23.5 & 28.0 \\
    LayerCAM \cite{jiang2021layercam}     & 17.69 & 23.5 & 17.9 \\
    PuzzleCAM \cite{jo2021puzzle}         & 33.52 & 24.8 & 45.7 \\
    MCTformer+-tiny \cite{xu2024mctformer+}           & 36.90 & 5.7  & 10.9 \\
    MCTformer+-small \cite{xu2024mctformer+}          & 31.13 & 21.6 & 11.0 \\
    \midrule
    \multicolumn{4}{l}{\textit{Temporal, external cues}}\\
    POF-CAM \cite{marelli2024temporal}    & 38.70 & 25.1 & 28 \\
    \midrule
    \multicolumn{4}{l}{\textit{Temporal, self-supervised (ours)}}\\
    MemCAM-tiny                              & \textbf{41.21} & 6.2  & 13.3 \\
    MemCAM-small                             & 25.03 & 26.4 & 13.4 \\
    \bottomrule
    \end{tabular}
    \caption{Comparison of mIoU, parameter count, and inference time per frame on WS$^2$ \cite{marelli2025ws2}. MemCAM achieves competitive results with substantially fewer parameters and faster inference.}
    \label{tab:miou_table_compact}
\end{table}

Tables~\ref{tab:ytov22_params} and~\ref{tab:miou_table_compact} show that MemCAM achieves state-of-the-art performance on both benchmarks, demonstrating its effectiveness in both object localization on YTOv2.2 and segmentation on WS$^2$, consistently outperforming the MCTformer+ baseline. This suggests that our cross-attention mechanism improves temporal consistency while requiring fewer parameters and achieving faster inference than the nearest competitors. Interestingly, on the large and diverse YTOv2.2 dataset (\(\sim\)720k frames), MemCAM scales effectively, with the Small variant performing strongly. However, on the much smaller and highly specific WS$^2$ dataset (\(\sim\)11k frames), the larger model variants (both MCTformer+-Small and MemCAM-Small) exhibit reduced performance. This behaviour suggests a capacity-data mismatch, consistent with  the intrinsic complexity of learning cross-attention under limited supervision. Learning temporal attention is a highly data-hungry optimization problem. As shown in Table~\ref{tab:miou_table_compact}, the larger teacher model (MCTformer+-Small) already overfits severely on this 11k-frame dataset, performing worse than its Tiny counterpart. Introducing  additional cross-attention weights on top of an oversized, overfitting model further exacerbates the issue, preventing meaningful temporal relationships from being learned. This observation is consistent with our ablation findings (Sec.~\ref{sec:ablation}), where cross-attention fails to converge when the supervisory signal is insufficient relative to the network's capacity. In contrast, the \textbf{MemCAM\_Tiny} variant achieves the best overall performance under this regime. This suggests that proper alignment between architectural capacity  and dataset scale is important for learning temporal attention under weak supervision. Further analyses, covering the effective temporal gain over teacher-student self-distillation, the propagated CAM used as the final prediction, classification accuracy and domain knowledge requirements, failure cases, and the applicability to other public benchmarks, can be found in the Supp., in Sec. ~\ref{subsec:supp_classification}, ~\ref{sec:supp_distill}, ~\ref{sec:supp_failure}.

\section{Conclusion}
We introduce MemCAM to successfully trains inter-frame cross-attention for WS-VSS. By replacing fragile external priors with learned semantic similarity in the latent space, CAMPA self-generates robust pseudo-labels that propagate class activations from past frames to the current one, providing the dense supervisory signal needed to optimize temporal weights. While learning cross-attention remains data-hungry, as it requires alignment between model capacity and dataset size, our feature-space matching approach is highly versatile. In future, we plan to extend this self-supervisory mechanism to other label-scarce settings such as unsupervised or semi-supervised Video Semantic Segmentation.

%% file: sec/X_suppl.tex
\clearpage
\setcounter{page}{1}
\maketitlesupplementary
\appendix

This document provides additional theoretical derivations, implementation details, and extended experiments that complement the main paper. The supplementary material is organized as follows. Sections~\ref{sec:mem_attn_details} and~\ref{sec:supp_campa} expand the theoretical formulation of the Memory Attention Module (MAM) and the CAM Propagation Algorithm (CAMPA), presenting details omitted from the main paper such as normalization choices, distributional assumptions, and the rationale behind the design of cross-frame attention and CAM diffusion. Section~\ref{sec:supp_training} explores in deep the description of the training and evaluation pipeline, including the sequential dataloading strategy, CAM generation process, temporal-window management, and additional considerations required for stable and efficient training of MemCAM. Finally, Sections~\ref{sec:ablation_supp} and~\ref{sec:supp_ws2} extend the ablation studies and dataset analysis presented in the main paper. Section \ref{sec:ablation_supp} reports further experiments on individual architectural components and training configurations, while Section \ref{sec:supp_ws2} provides an expanded discussion of the characteristics and challenges of the WS$^2$ dataset and their impact on model behavior. Finally, Section~\ref{sec:supp_distill} disentangles the temporal contribution of MemCAM from teacher--student self-distillation and discusses the correspondences it learns, Section~\ref{sec:supp_scope} details the backbone requirements that CAMPA imposes and delimits the scope of MemCAM, while Section~\ref{sec:supp_failure} analyzes its failure cases. Together, these additions offer a more comprehensive and transparent account of the methodological and experimental choices underlying MemCAM.

\begin{figure*}[t]
    \centering
    \includegraphics[width=0.9\linewidth]{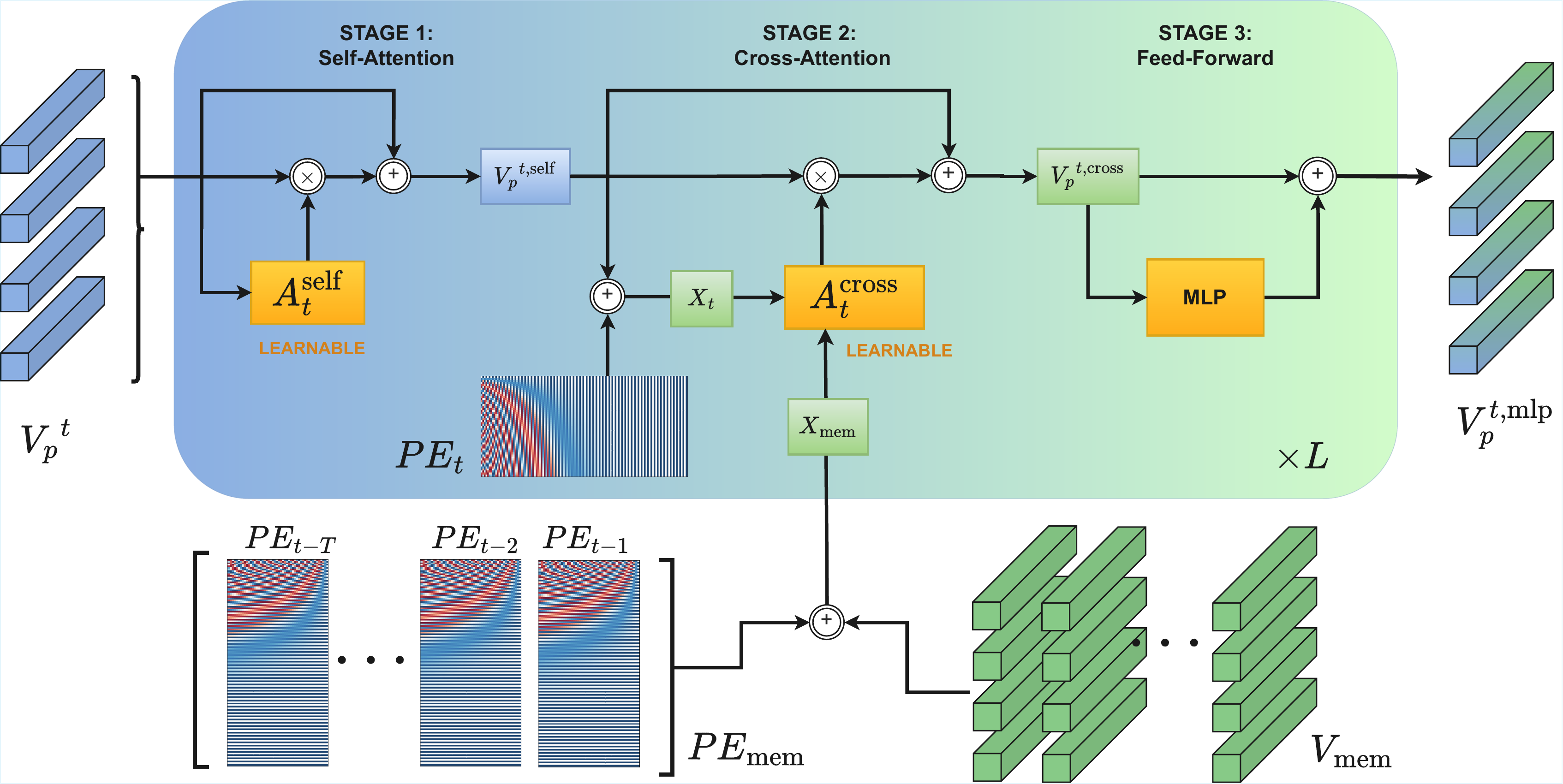}\caption{
    \textbf{Detailed architecture of the MAM}. Starting from the backbone patch embeddings $V_p^t$, the module refines representations through three stages, each equipped with a residual connection.
    \textbf{Self-attention (left)}: current-frame tokens act as queries, keys, and values. The arrows entering the yellow block indicate that the attention matrix $A_t^{\mathrm{self}}$ is \emph{learned} from the input; its output $\mathrm{Attn}(\mathbf{q}^{\mathrm{self}},\mathbf{k}^{\mathrm{self}},\mathbf{v}^{\mathrm{self}})$ is added back to the backbone features, producing the self-attended representation $V_p^{t,\mathrm{self}} = V_p^t + \mathrm{Attn}(\mathbf{q}^{\mathrm{self}},\mathbf{k}^{\mathrm{self}},\mathbf{v}^{\mathrm{self}})$.
    \textbf{Cross-attention (center)}: the refined tokens $V_p^{t,\mathrm{self}}$ query the memory tokens stored in the patch queue $\mathcal{Q}_p$, yielding a learned cross-attention matrix $A_t^{\mathrm{cross}}$. Absolute sinusoidal temporal positional encodings $PE_t$ and $PE_{\mathrm{mem}}$ are added respectively to $V_p^{t,\mathrm{self}}$ and to the memory tokens to distinguish frames in time. The resulting cross-attention injects temporal information from past frames and updates the representation via the skip connection $V_p^{t,\mathrm{cross}} = V_p^{t,\mathrm{self}} + \mathrm{Attn}(\mathbf{q}^{\mathrm{cross}},\mathbf{k}^{\mathrm{cross}},\mathbf{v}^{\mathrm{cross}})$.
    \textbf{Feed-forward block (right)}: a two-layer MLP further refines each token independently. Its output is added through a residual connection to obtain ${V_p^t}' = V_p^{t,\mathrm{cross}} + \mathrm{MLP}(V_p^{t,\mathrm{cross}})$, which constitutes the final temporally enriched patch embedding carrying both spatial and temporal context.}

    \label{fig:memory_attention}
\end{figure*}

\section{Additional details on the Memory Attention Module}
\label{sec:mem_attn_details}

The \textbf{Memory Attention Module (MAM)} introduced in the main paper (Section~\ref{sec:mem_attn}) refines backbone patch embeddings by injecting temporal context stored in the \emph{Memory Bank}. Its design draws inspiration from the decoder-like memory mechanism of SAM~2~\cite{ravi2024sam} and from the Transformer architecture of~\cite{vaswani2017attention}.
The detailed internal structure of the block is illustrated in Figure~\ref{fig:memory_attention}, which
 visually summarizes the full process.

\paragraph{MAM overview}

The MAM receives backbone patch embeddings of the current frame,
\[
V_p^t \in \mathbb{R}^{N^2 \times D},
\]
and refines them by incorporating information stored in a Memory Bank containing the previous \(T\) frames.
This refinement proceeds through three sequential stages:
(1) self-attention within the current frame,
(2) cross-attention with past-frame tokens stored in memory,
(3) a feed-forward refinement block.

Each stage transforms the representation in a controlled manner, and each transformation is regulated by residual connections, allowing the network to smoothly blend new temporal information with the original backbone features.
By the end of the block, the resulting embedding \(V_p^{t'}\) contains both local spatial structure and temporal context.

\paragraph{Attention Preliminaries}

As in standard Transformers~\cite{vaswani2017attention}, attention operates on queries, keys, and values:
\[
\mathbf{q} = X W_q,\quad
\mathbf{k} = X W_k,\quad
\mathbf{v} = X W_v,
\]
and computes:
\begin{equation}
\label{eq:supp_attention}
\mathrm{Attn}(\mathbf{q},\mathbf{k},\mathbf{v})
=
\mathrm{softmax}\!\left(
    \frac{\mathbf{q}\mathbf{k}^{\top}}{\sqrt{D}}
\right)\mathbf{v}.
\end{equation}

Intuitively, attention creates a weighted average of “value” vectors, where weights indicate semantic similarity between “query” and “key” vectors.
In our case, self-attention explores spatial locations within the same frame, whereas cross-attention discovers correspondences between the current frame and the memory of past frames.

\paragraph{Stage 1: Self-Attention Within the Current Frame}

The process begins by letting all patches in the current frame exchange information with each other.
This allows the model to refine local details and capture global spatial structure, without yet using memory.

We form:
\[
\mathbf{q}^{\mathrm{self}},\;
\mathbf{k}^{\mathrm{self}},\;
\mathbf{v}^{\mathrm{self}}
=
V_p^t W_{q,k,v}^{\mathrm{self}}
\]
and compute:
\[
A_t^{\mathrm{self}}
=
\mathrm{Attn}(
\mathbf{q}^{\mathrm{self}},
\mathbf{k}^{\mathrm{self}},
\mathbf{v}^{\mathrm{self}}
).
\]

Each patch thus receives information from every other patch within the same frame.
This is particularly important for weakly supervised settings, where no pixel-wise annotations guide localization: self-attention implicitly groups spatial patterns and enhances structural coherence.

A residual update gives:
\[
V_p^{t,\mathrm{self}}
=
V_p^t + A_t^{\mathrm{self}}.
\]

\paragraph{Stage 2: Cross-Attention With the Memory Bank}

The Memory Bank stores the patch embeddings of the previous \(T\) frames:
\[
\mathcal{Q}_p
=
[\, V_p^{t-T}, \dots, V_p^{t-1} \,]
\in \mathbb{R}^{T \times N^2 \times D}.
\]

To allow the model to distinguish where each token comes from in time, we inject absolute sinusoidal positional encodings.
For a frame at time index \(t\), the encoding is:
\[
PE_t(2i)   = \sin\!\left(t/10000^{\frac{2i}{D}}\right),
\]
\[
PE_t(2i+1) = \cos\!\left(t/10000^{\frac{2i}{D}}\right),
\]
which is then broadcast across all spatial patches, yielding \(PE_t \in \mathbb{R}^{N^2 \times D}\).

Past-frame positional encodings are stacked into one large tensor:
\[
PE_{\mathrm{mem}}
=
    [\,PE_{t-T}, \dots, PE_{t-1}\,]
    \in \mathbb{R}^{T N^2 \times D}.
\]

These encodings play a crucial role: they endow attention with the ability to reason not only over which patches are semantically similar but also \emph{when} they occurred.

After the self-attention step, the model explicitly interacts with the temporal dimension.
The refined current tokens query the entire memory of past patches, allowing the network to retrieve contextual information that might no longer be visible in the current frame (e.g., due to occlusion or motion blur).

Current tokens receive positional encoding:
\[
X_t = V_p^{t,\mathrm{self}} + PE_t.
\]

Memory tokens are flattened and encoded similarly:
\[
V_{\mathrm{mem}} = \mathrm{reshape}(\mathcal{Q}_p),
\quad
X_{\mathrm{mem}} = V_{\mathrm{mem}} + PE_{\mathrm{mem}}.
\]

Projections follow:
\[
\mathbf{q}^{\mathrm{cross}} = X_t W_q^{\mathrm{cross}},
\qquad
\mathbf{k}^{\mathrm{cross}} = X_{\mathrm{mem}} W_k^{\mathrm{cross}},
\]
\[
\mathbf{v}^{\mathrm{cross}} = V_{\mathrm{mem}} W_v^{\mathrm{cross}}.
\]

Then attention is computed:
\[
A_t^{\mathrm{cross}}
=
\mathrm{Attn}\!\left(
\mathbf{q}^{\mathrm{cross}},
\mathbf{k}^{\mathrm{cross}},
\mathbf{v}^{\mathrm{cross}}
\right).
\]

This operation can be interpreted as:
For every spatial location in the current frame, find the most similar spatial locations across the past frames, taking into account both visual similarity and temporal ordering.”

The updated representation becomes:
\[
V_p^{t,\mathrm{cross}}
=
V_p^{t,\mathrm{self}} + A_t^{\mathrm{cross}}.
\]

\paragraph{Stage 3: Feed-Forward Refinement}

The third stage acts as a point-wise transformation.
It refines the representation after the temporal retrieval and adjusts the scale and distribution of features.

A two-layer MLP with GELU activation:
\[
\mathrm{MLP}(x)
=
W_2 \,\mathrm{GELU}(W_1 x + b_1) + b_2,
\]
is applied to each patch independently.

Another residual connection produces:
\[
V_p^{t,\mathrm{mlp}}
=
V_p^{t,\mathrm{cross}}
+
\mathrm{MLP}(V_p^{t,\mathrm{cross}}).
\]

Finally, the output of the entire MAM is obtained by blending the temporal refinement with the original backbone output:
\begin{equation}
\label{eq:supp_final_residual}
V_p^{t'}
=
V_p^t + V_p^{t,\mathrm{mlp}}
\quad\in\mathbb{R}^{N^2 \times D}.
\end{equation}

This dual-residual approach stabilizes training and prevents the model from overfitting to memory content in situations where temporal information may be unreliable.

\subsection{Interpretation and Intuition}

The overall block can be interpreted as follows:

\begin{itemize}
    \item \textbf{Self-attention} provides the model with a clean and spatially coherent description of the current frame.
    \item \textbf{Cross-attention} injects temporal cues by retrieving visually and semantically aligned patches from the past.
    \item \textbf{The feed-forward network} consolidates these signals into a richer embedding.
    \item \textbf{The final residual connection} allows the model to decide how much temporal information should influence the final representation.
\end{itemize}

This makes MAM particularly well-suited for weak supervision, where explicit annotations do not exist and where temporal coherence becomes a strong implicit supervisory signal.

\subsection{Relation to SAM 2}

Although the conceptual idea of storing past embeddings resembles SAM~2~\cite{ravi2024sam}, the purposes and constraints are entirely different:

\begin{itemize}
    \item SAM~2 uses a complex \textbf{Memory Encoder} to fuse masks, object pointers, and compressed visual tokens; in contrast, MemCAM stores raw patch embeddings and CAMs to avoid adding supervision-dependent components.
    \item SAM~2 relies on Rotary Positional Embeddings; we adopt simpler sinusoidal temporal encodings because they generalize well without additional supervision.
    \item Our memory design is significantly lighter, avoiding the 10--20\% parameter increase of SAM~2, which would be prohibitive in weakly supervised training.
\end{itemize}

Thus, while inspired by SAM~2's architectural philosophy, MAM is redesigned from scratch to suit the weakly supervised, mask-free learning scenario of video segmentation.

\section{Additional details on CAM Propagation Algorithm}
\label{sec:supp_campa}

This section provides an extended description of the \textbf{CAM Propagation Algorithm} (CAMPA), complementing the main paper (Section~\ref{sec:cam_loss}).
Here we detail all intermediate transformations, normalization steps, and matrix operations that produce the synthetic, temporally propagated CAM $\hat{M}^t$.
The complete process is illustrated in Figure~\ref{fig:cam_propagation_algorithm}.

\paragraph{CAMPA overview}

CAMPA generates a \emph{self-supervisory pseudo-label} for each current frame $I_t$ by transferring CAM activations from the previous $T$ frames, according to the patch-level feature similarity computed in the shared embedding space.
This synthetic CAM behaves as a temporally coherent target and is used to supervise the predicted CAM through the consistency loss described later.

Unlike classical temporal smoothing methods, CAMPA relies exclusively on \textbf{feature-space correspondence}.
It does not assume small motion, constant color, or smooth trajectories, and does not require external cues such as optical flow.

Algorithm~\ref{alg:campa} summarizes the whole procedure; the paragraphs below detail each step.

\vspace{0.2cm}
\begin{algorithm}[h]
\caption{CAM Propagation Algorithm (CAMPA)}
\label{alg:campa}
\begin{algorithmic}[1]
\State \textbf{Input:} current patch embeddings $V_p^t\in\mathbb{R}^{N^2\times D}$; patch queue $\mathcal{Q}_p=[V_p^{t-T},\dots,V_p^{t-1}]$; CAM queue $\mathcal{Q}_M=[M_c^{t-T},\dots,M_c^{t-1}]$; decay $\lambda\in(0,1]$; sparsification $K$; temperature $\mu$
\State \textbf{Output:} propagated CAM $\hat{M}^t\in\mathbb{R}^{N^2}$
\State $\widehat{V}_p^{\,t} \gets \ell_2\text{-normalize rows of } V_p^t$
\For{$\tau = 1$ to $T$} \Comment{most recent frame first}
    \State $\widehat{V}_p^{\,t-\tau} \gets \ell_2\text{-normalize rows of } V_p^{t-\tau}$
    \State $s_{t-\tau} \gets \widehat{V}_p^{\,t}\,(\widehat{V}_p^{\,t-\tau})^{\top}$ \Comment{cosine similarity, $N^2\!\times\!N^2$}
    \State $\widetilde{s}_{t-\tau} \gets \lambda^{\tau}\, s_{t-\tau}$ \Comment{temporal decay}
\EndFor
\State $S \gets [\,\widetilde{s}_{t-1}\,|\,\widetilde{s}_{t-2}\,|\dots|\,\widetilde{s}_{t-T}\,]\in\mathbb{R}^{N^2\times(T N^2)}$
\For{each row $i$ of $S$}
    \State $S'(i,j) \gets S(i,j)$ if $S(i,j)$ is among the Top-$K$ values of $S(i,:)$, else $-\infty$
\EndFor
\State $P \gets \text{row-wise softmax}\big(S'/\mu\big)$ \Comment{$N^2\!\times\!(T N^2)$, non-Top-$K$ entries get zero mass}
\State $\hat{M}^t(i) \gets \sum_{\tau=1}^{T}\sum_{j=1}^{N^2} P\!\left(i,(\tau-1)N^2+j\right) M_c^{\,t-\tau}(j)$
\State \Return $\hat{M}^t$
\end{algorithmic}
\end{algorithm}

\subsection{Building the probability Matrix}
\paragraph{Step 1: Feature Normalization}

Let $V_p^t \in \mathbb{R}^{N^2 \times D}$ be the current-frame patch embeddings and
$\mathcal{Q}_p = [V_p^{t-T}, \dots, V_p^{t-1}]$ the patch memory queue (see Eq.~\ref{eq:patch_queue}).
For each frame $t-\tau$, we normalize every patch embedding using row-wise $\ell_2$ normalization:
\[
\widehat{V}_{p}^{\,t-\tau}(i,:) =
\frac{V_{p}^{\,t-\tau}(i,:)}{\|V_{p}^{\,t-\tau}(i,:)\|_2 + \varepsilon},
\qquad i = 1,\dots,N^2,
\]
where $V_{p}^{\,t-\tau}(i,:) \in \mathbb{R}^D$ represent the feature space of a single patch of the frame $I_{t-\tau}$ and  $\varepsilon$ ensures numerical stability. The same normalization is applied for the feature space $V_p^t$ of the current frame.

\paragraph{Step 2: Cosine Similarity and Temporal Decay}

As in Eq.~(\ref{eq:comp_sim_submatrixes}) of the main paper, cosine similarities between current and past patches are computed frame by frame:
\[
s_{t-\tau}
=
\widehat{V}_p^{\,t}
\left(\widehat{V}_p^{\,t-\tau}\right)^{\!\top}
\in\mathbb{R}^{N^2\times N^2}.
\]

To emphasize recent frames, each similarity matrix is multiplied by the temporal decay factor $\lambda^\tau$ (with $\lambda\in (0,1]$):
\begin{equation}
    \label{eq:time_decay}
\widetilde{s}_{t-\tau} = \lambda^\tau s_{t-\tau}.
\end{equation}
\paragraph{Step 3: Block Concatenation of Temporal Similarities}

We then concatenate all decayed similarities into a single matrix:
\[
S
=
\left[
\widetilde{s}_{t-1}\,
\Big|\,
\widetilde{s}_{t-2}\,
\Big|\,
\dots
\Big|\,
\widetilde{s}_{t-T}
\right]
\in\mathbb{R}^{N^2 \times (T N^2)}.
\]

Each row of $S$ lists all similarity scores associated with a \emph{single patch of the current frame} versus \emph{every patch of the past $T$ frames}.

\paragraph{Step 4: Row-wise Top-$K$ Filtering}

Similarity matrices are dense and contain many weak or noisy matches.
To isolate reliable correspondences, we keep only the $K$ largest values in each row.

Formally, for each row vector $S(i,:)$ we define:
\begin{equation}
\label{eq:using_topK}
S'(i,j)
=
\begin{cases}
S(i,j), & \text{if } S(i,j) \text{ in Top-$K$ values of row } i, \\
-\infty, & \text{otherwise}.
\end{cases}
\end{equation}
This operation removes spurious matches, sparsifies the similarity graph and ensures the softmax will focus on $K$ meaningful correspondences only.

The result is $S' \in \mathbb{R}^{N^2 \times (TN^2)}$, a sparse version of $S$.

\paragraph{Step 5: Temperature-Scaled Softmax}

To convert similarities into a valid probability distribution, we apply a row-wise softmax with optional temperature parameter $\mu > 0$:

\begin{equation}
\label{eq:using_temperature}
P(i,j)
=
\frac{
\exp\!\left( S'(i,j) / \mu \right)
}{
\sum_{k=1}^{TN^2}
\exp\!\left( S'(i,k) / \mu \right)
}.
\end{equation}

Because non–Top-$K$ entries were set to $-\infty$, they contribute zero probability.
The output matrix,

\[
P \in [0,1]^{N^2 \times (TN^2)},
\]

assigns, for each patch of the current frame, a distribution over all patches of all past frames.

\subsection{CAM Supervision}
\paragraph{Step 6: CAM Propagation}

Past class-specific CAMs stored in the CAM queue
$\mathcal{Q}_M = [M_c^{t-T}, \dots, M_c^{t-1}]$
are flattened and aligned with the block structure of $P$.

The propagated CAM, as in Eq.~(\ref{eq:propagated_cam}) of the main paper, is obtained via:

\[
\hat{M}^t(i)
=
\sum_{\tau=1}^{T}
\sum_{j=1}^{N^2}
P\!\left(i,\,(\tau-1)N^2 + j\right)
\,M_c^{\,t-\tau}(j).
\]

Thus, patch $i$ in the current frame receives a weighted combination of all matching patches from previous frames.

\paragraph{Step 7: Normalization Before the Consistency Loss}

Both synthetic and real maps must represent probability distributions.
Thus we apply $\ell_1$ normalization:

\[
\hat{M}^{t\,\prime}
=
\frac{\hat{M}^t}{\|\hat{M}^t\|_1},
\qquad
{M_c^{t}}^{\!\prime}
=
\frac{M_c^t}{\|M_c^t\|_1}.
\]

Now:
\[
\hat{M}^{t\,\prime},\,{M_c^{t}}^{\!\prime}
\in
\Delta^{N^2 - 1},
\]
This normalization ensures that both CAMs lie in the $(N^2-1)$-dimensional probability simplex ($\Delta^{N^2-1}$), meaning that it can be interpreted as a probability distribution over the set of patches.

\paragraph{Step 8: KL-Divergence Consistency Loss}

The final loss term encourages the predicted CAM to approach the temporally propagated one.
Using the normalized distributions, the consistency loss is:

\[
\mathcal{L}_{\mathrm{M}}
=
D_{\mathrm{KL}}
\!\left(
\hat{M}^{t\,\prime}
\;\middle\|\;
{M_c^{t}}^{\!\prime}
\right)
=
\sum_{i=1}^{N^2}
\hat{M}^{t\,\prime}(i)
\log
\frac{
\hat{M}^{t\,\prime}(i) + \varepsilon
}{
{M_c^{t}}^{\!\prime}(i) + \varepsilon
}.
\]

This penalizes discrepancies between the predicted activation pattern and the synthetic, temporally propagated target.

\subsection{Interpretation}

CAMPA behaves as a \emph{temporal teacher}:
the model reuses its past CAMs to supervise future ones.
If previous frames identify the object reliably, this information is propagated forward; if not, the network still gradually stabilizes thanks to repeated similarity-based averaging. The algorithm also implicitly corrects local CAM failures: a noisy activation in the current frame will be pulled toward the more reliable consensus of past frames. Through this self-supervised mechanism, temporal coherence is learned without dense masks, external motion estimation, or handcrafted priors. The edge case in which the entire student CAM is zero is handled in Sec.~\ref{sec:training_stability}.

\begin{figure*}[t]
    \centering
    \includegraphics[width=1.0\linewidth]{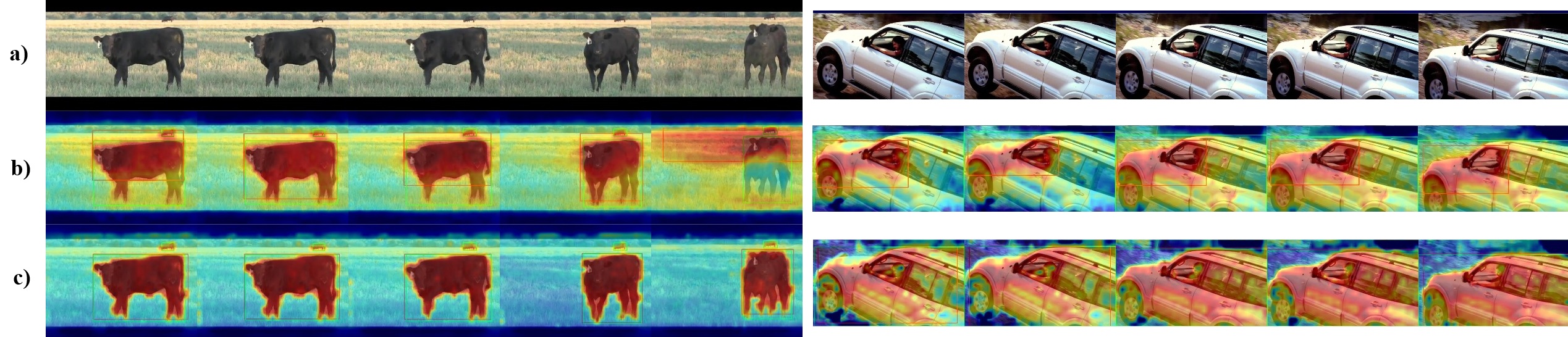}
    \caption{Comparison of temporal consistency between \textbf{Ground Truth (a)} \textbf{MCTformer+\_tiny (b)} and \textbf{MemCAM\_tiny (c)}.
    Each sequence shows consecutive frames from the same clip.}
    \label{fig:qualitative_comparison_seq}
\end{figure*}

\section{Training Strategy and Implementation Details}
\label{sec:supp_training}

This section provides additional details on the training strategy used by MemCAM, complementing Section~\ref{sec:experiments} of the main paper.
Our model requires sequential input because the MAM relies on a fully populated memory buffer containing the past $T$ frames.
Several challenges arise from this requirement: (i) processing sequential frames increases overfitting risk; (ii) storing representations produced with outdated parameters causes feature mismatch; (iii) training-time noise degrades the quality of stored patch embeddings; (iv) standard dataloaders break temporal coherence due to inconsistent augmentations; (v) sequential per-frame processing reduces GPU parallelization; and (vi) the CAM refinement pipeline is not designed for training and may destabilize gradients.
We address all these challenges through a dedicated set of dataloaders, an augmentation strategy tailored to video clips, and a Teacher–Student training protocol described below.

\paragraph{Efficient Sequential Dataloaders.}
MemCAM requires frames to be processed in strict temporal order so that each frame can retrieve patch tokens and CAMs from the previous $T$ frames stored in the Memory Bank. We therefore introduce two custom dataloaders:

\textbf{SeqDataloader} (training).
Given a video of length $T_{\text{video}}$, the dataloader first divides it into non-overlapping windows of length $L_w$, where $L_w$ is a multiple of the subclip length $T$.
Within each window, frames are randomly partitioned into $L_w/T$ subclips of length $T$.
This sampling strategy reduces redundancy (particularly in high-FPS videos) and enables learning temporal relationships at different scales depending on the choice of $(L_w,T)$.
A random sampler then groups these subclips into batches of shape $T \times 3 \times H \times W$, allowing full GPU parallelization while preserving sequential structure.

\textbf{SeqDataloaderMS} (multi-scale inference).
To avoid redundant forward passes and memory contamination during multi-scale evaluation, this dataloader returns, for each scale, a tensor containing both the original and horizontally flipped versions of the frame. Each scale is processed by a dedicated model copy, preventing interference between memory buffers at different resolutions and enabling parallel inference.

\paragraph{Clip-Consistent Data Augmentation.}
To maintain temporal coherence during training, augmentations are sampled once per subclip and applied consistently across all $T$ frames.
We use horizontal flips, zoom-in cropping, grayscale conversion, color jittering (brightness, contrast, saturation, hue), and random erasure.
This preserves motion-consistency while providing regularization and visual diversity.

\paragraph{Teacher–Student Training Protocol.}
To ensure stable memory representations and avoid the mismatch introduced by updating parameters within a sequence, we employ a Teacher–Student scheme (Algorithm~\ref{alg:training_protocol}).
A frozen MCTformer+ model \cite{xu2024mctformer+} serves as Teacher, responsible for generating the patch embeddings and CAMs used to populate the Memory Bank for frames $1,\dots,T-1$.
MemCAM (the Student) processes only the final frame $T$, retrieving memory filled entirely with Teacher-generated features.
A synthetic CAM for the final frame is then produced using the CAM Propagation Algorithm (Section~\ref{sec:cam_loss}), and the Student is optimized using the classification loss and the CAM Consistency Loss.

\vspace{0.2cm}
\begin{algorithm}[h]
\caption{Training Protocol}
\label{alg:training_protocol}
\begin{algorithmic}[1]
\State \textbf{Input:} Teacher \textbf{Teach}, Student \textbf{Stud}, subclip $J \in \mathbb{R}^{T \times 3 \times H \times W}$
\State \textbf{Teach} $\gets$ eval mode
\State \textbf{Stud} $\gets$ train mode
\State \textbf{Stud.Memory.reset}()
\For{$t = 1$ to $T - 1$}
    \State patch\_tokens ($V_p^t$), CAMs (${V_M^t}'$) $\gets$ \textbf{Teach}($J[:, t]$)
    \State \textbf{Stud.Memory.add}(patch\_tokens, CAMs)
\EndFor
\State final\_tokens, $_{}$ $\gets$ \textbf{Teach}($J[:, T]$)
\State $ \hat{M^t} \gets \textbf{Stud.Memory.get\_target\_cams}(final\_tokens)$
\State output $\gets$ \textbf{Stud}($J[:, T]$)
\State compute loss $\mathcal{L}(\text{output}, \text{targets}, \hat{M^t})$
\State backprop and update \textbf{Stud}
\end{algorithmic}
\end{algorithm}

This protocol resolves (i)–(iii): the Student is never trained on its own noisy intermediate representations; the memory is always consistent; and frames remain sequential only within subclips, while sampling introduces stochasticity across updates.

\paragraph{Multi-Scale CAM Generation.}
For inference (Algorithm~\ref{alg:inference_protocol}), we follow a multi-scale strategy similar to~\cite{xu2024mctformer+}, but improved for temporal models.
Each scale is processed by an isolated Student model instance, preventing cross-scale contamination of the memory queue.
The patch embeddings and CAMs are computed sequentially for each frame, and CAMs from all scales are merged before refinement.

\vspace{0.2cm}
\begin{algorithm}[h]
\caption{CAM Generation Protocol}
\label{alg:inference_protocol}
\begin{algorithmic}[1]
\State \textbf{Input:} Student copies $[\text{Stud}_{s_1}, \dots, \text{Stud}_{s_n}]$, scale inputs $J=\{ J_{s_1}, \dots, J_{s_n} \}$
\For{each scale $s$}
    \State set $\text{Stud}_{s}$ to eval mode and reset memory
\EndFor
\State FINAL\_CAMS $\gets$ [ ]
\For{$t = 1$ to $T$}
    \State CAMS\_tensors $\gets$ [ ]
    \For{each scale $s$}
        \State tokens, $\text{CAMs}_{s}$ $\gets$ $\text{Stud}_{s}\big(J_{s}[t]\big)$
        \State $\text{Stud}_{s}.\text{Memory.add}(\text{tokens},\, \_ )$
        \State append $\text{CAMs}_{s}$ to CAMS\_tensors
    \EndFor
    \State final\_cams $=$ merge\_cams(CAMS\_tensors)
    \State append final\_cams to FINAL\_CAMS
\EndFor
\State \Return FINAL\_CAMS
\end{algorithmic}
\end{algorithm}

This inference protocol ensures efficient multi-scale processing while preserving temporal consistency across all scales.

\subsection{Training Stability}
\label{sec:training_stability}
Adding a CAM Consistency Loss meant extending the problem from classification-only to also including a KL-divergence minimization. This introduced some theoretical challenges we had to solve.

\paragraph{CAM Normalization.}
In the previous sections, we introduced the CAM Consistency Loss and emphasized the importance of applying $\ell_1$ normalization.
This choice is different from the one made in \cite{xu2024mctformer+}, where the authors opt for a \textit{minmax} normalization.
While \textit{minmax} is a viable option when CAMs are generated in evaluation mode and not subject to gradient flow, it is undesirable during training, as it allows the model to modify the final CAM and change the loss values just by changing the minima and maxima.
On the other hand, $\ell_1$ normalization introduces a probabilistic interpretation of CAMs and forces the model to actually move the CAMs' mass during learning, not just its extrema.

\begin{figure*}[t]
    \centering
    \begin{subfigure}[t]{0.48\textwidth}
        \centering
        \includegraphics[width=\textwidth,height=0.75\textwidth,keepaspectratio=false]{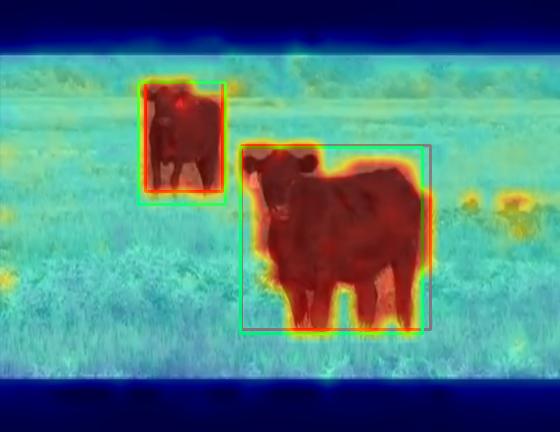}
        \caption{Image with CorLoc = 1}
        \label{fig:sub1}
    \end{subfigure}
    \hfill
    \begin{subfigure}[t]{0.48\textwidth}
        \centering
        \includegraphics[width=\textwidth,height=0.75\textwidth,keepaspectratio=false]{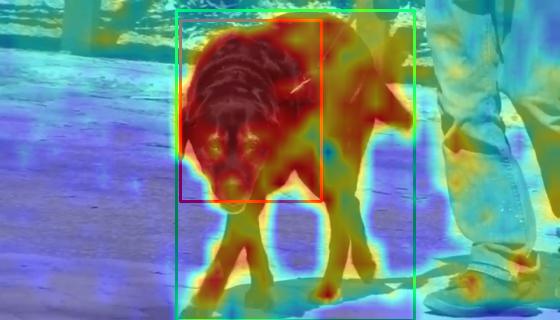}
        \caption{Image with CorLoc = 0}
        \label{fig:sub2}
    \end{subfigure}

    \caption{Comparison of CorLoc scores for two examples.
Green bounding boxes denote ground-truth annotations, while red boxes indicate model predictions.
Figure~\ref{fig:sub1} shows an almost perfect overlap between the predicted and ground-truth boxes, whereas Figure~\ref{fig:sub2} illustrates a poor localization result.}
    \label{fig:corloc_comparison}
\end{figure*}

\paragraph{CAM Compression.}
During the CAM refinement procedure (Section~\ref{sec:patch_supervision}), the authors of MCTformer+~\cite{xu2024mctformer+} introduced a square root operation to compress the range of the CAMs obtained by multiplying the class-to-patch attentions weights $A_{c2p}$ \cite{xu2024mctformer+} with the CONV head outputs $V_M^t$.
In MCTformer+, this worked as intended because CAMs were generated only during evaluation, and no gradient was backpropagated through this step.  However, in our setting CAMs actively participate in the loss computation.
As a result, \texttt{torch} includes them in the computational graph, and gradients are propagated through the square root operation:
\[
\frac{d}{dx} \sqrt{x} = \frac{1}{2\sqrt{x}}.
\]

This derivative leads to two problematic situations:
low activation regions cause exploding gradients, since $\lim_{x \to 0^+} \frac{1}{2\sqrt{x}} = +\infty$,
while very high activations lead to vanishing gradients, as $\lim_{x \to +\infty} \frac{1}{2\sqrt{x}} = 0$.
It is important to remark that \cite{xu2024mctformer+} introduced the square root as a compression function, therefore the relative distribution of the CAM remains unchanged.
Additionally, because the square root is monotonic, minimizing $\sqrt{x}$ is equivalent to minimizing $x$ itself.
Therefore, during training we omit the compression step and redefine the refinement rule as:
\[
{V^{t}_{M}}'(c,i,j) = A_{c2p}(c,i,j) \cdot V^{t}_{M}(c,i,j).
\]
where $A_{c2p}$ is defined as in \cite{xu2024mctformer+}.
\paragraph{Row-Stochastic Attention}
In \cite{xu2024mctformer+} the authors further refine CAMs using patch-to-patch attention weights $A_{p2p}$.
However, they do not enforce that the attention weight matrix is row-stochastic. This omission has two main consequences.

(i) From a CAM generation perspective, patches with low activation can appear highly activated after refinement. Since attention weights are not normalized, each resulting patch activation is not a convex combination of past activation, but rather a weighted sum. Therefore, patches with uniformly high attention weights towards all patches will exhibit high activations, regardless of their actual semantic relevance.

(ii) From a training perspective, this uncertainty introduces instability. During training the model has to account for the unbounded and unpredictable magnitudes of attention weights. On the other hand, enforcing row normalization ensures stability, making each refined patch's activation dependent only on the \emph{relative} strength of its attention weights, not their magnitude.

We thus normalize the attention matrix to make it row-stochastic before refinement.
The new attention matrix is computed as:

\[
D(i,i) = \sum_{j=1}^{N} A_{p2p}(i,j),
\qquad
\hat{A}_{p2p} = D_b^{-1} A_{p2p}\]
and the refined CAM becomes:
\[
{V}^{t,\text{refined}}_{M}(c) = \hat{A}_{p2p} \, {V^{t}_{M}}'(c).
\]

where ${A}_{p2p}$ is defined as in \cite{xu2024mctformer+}.

\paragraph{KL Divergence}
Another issue with KL Divergence computation arose because of CAM entries that are equal to zero. In that case, two scenarios were possible:
\begin{enumerate}
    \item $q_i = 0$ (i.e. the teacher assigns zero probability to patch $i$). In this case, both the theoretical convention and $\texttt{torch}$'s implementation return $q_i\log(p_i) = 0$, which is the desired result, since it means that the loss will not penalize the Student because the Teacher assigns no probability to the patch.
    \item $p_i = 0$ and $q_i \neq 0$ (i.e. the teacher assigns nonzero probability to patch $i$, but the student assigns zero probability to it). In this case, we'd have $q_i\log(p_i) = \infty$, which is problematic. To fix the problem, we set the minimum value that $p_i$ can reach to $\varepsilon = 10^{-6}$, so that no $p_i$ ever equals 0.
\end{enumerate}

A particularly challenging case occurs when the entire student CAM is zero.
In this situation, $\ell_1$ normalization leaves the CAM unchanged, but the resulting vector no longer lies on the probability simplex, violating the assumption that CAMs represent valid probability distributions.
When in this situation, we choose to interpret the Student prediction as being maximally uncertain. We therefore replace its produced CAM with one containing all equal values, mimicking a uniform distribution over patches.
Formally, given $M = [0, \dots, 0] \in \mathbb{R}^N$, we define
\[
\hat{M} = \left[\tfrac{1}{N}, \dots, \tfrac{1}{N}\right] \in \Delta^{N-1},
\]
which allows the KL divergence to be computed while ensuring that the student CAM reflects maximal uncertainty.

\begin{figure*}[t]
    \centering
    \includegraphics[width=0.9\linewidth]{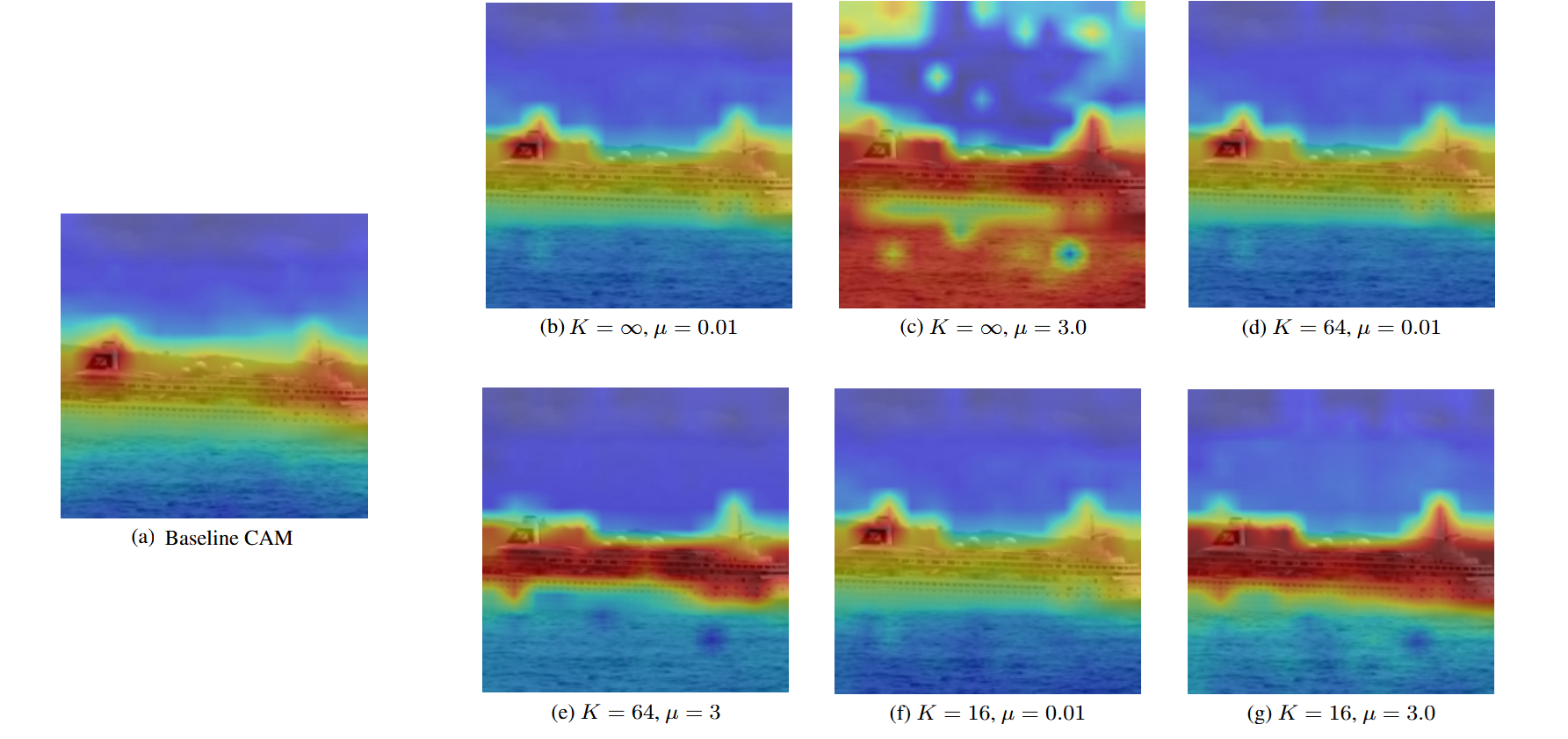}
    \caption{one sample frame with \(\lambda=0.97\) and various \((K,\mu)\). Panel (a) is the baseline \(\mathrm{CAM}_6\); panels (b, c, d, e, f, g) depict \(\mathrm{CAM}^{\mathrm{SUP}}\). A qualitative inspection shows that increasing \(\mu\) while decreasing \(K\) produces supervisory CAMs that are more uniform, significantly improving over the baseline in (a).}
    \label{fig:ablations_propagation}
\end{figure*}

\subsection{Evaluation Strategy}

\paragraph{Evaluation Metric.}  In order to benchmark our method against previous results in weakly supervised video object localization, such as \cite{belharbi2025colo} and \cite{belharbi2023tcam}, we adopt as evaluation score the \textbf{Correct Localization (CorLoc)} metric.
CorLoc measures the percentage of images in which the method correctly localizes at least one object of interest. A localization is considered correct if the IoU between the predicted bounding box and the ground truth bounding box is above a threshold $\tau$ (in previous works, $\tau = 0.5$)

\noindent Formally, let
\begin{itemize}
    \item $N$ denote the total number of images in the evaluation set,
    \item $\mathcal{G}_i$ the set of ground-truth bounding boxes in image $i$,
    \item $\hat{B}_i$ the predicted bounding box for image $i$,
    \item $\operatorname{IoU}(B, G) = \frac{|B \cap G|}{|B \cup G|}$ the IoU between bounding boxes $B$ and $G$.
\end{itemize}
We define the success indicator for image $i$ as
\[
s_i =
\begin{cases}
1, & \exists \, G \in \mathcal{G}_i \ \text{such that} \ \operatorname{IoU}(\hat{B}_i, G) \geq \tau, \\
0, & \text{otherwise}.
\end{cases}
\]
The CorLoc metric is then computed as
\[
\operatorname{CorLoc} = \frac{1}{N} \sum_{i=1}^{N} s_i.
\]

Clearly, a higher CorLoc indicates that a higher fraction of the total samples had one instance correctly localized by the model, which in turn implies a better localization capability. In order to perform a comprehensive evaluation of our method, we report both the overall CorLoc and the per-class CorLoc. This allows us to highlight classes on which the model performs particularly well or poorly. We show an example of good and bad localization, as well as the relative CorLoc scores, in Figure \ref{fig:corloc_comparison}.

\paragraph{Evaluation Protocol.}
The train-test split is provided by the authors of YouTube-Objects v2.2 and is kept unchanged.
For validation, we perform a 70/30 split of the training videos, carefully ensuring that no frames from the same video end up in different splits.

\paragraph{Mask and Bounding Box Extraction.}
\label{sec: bbox_extraction}
Since produced CAMs can be spread out and noisy, we apply additional geometric refinement techniques to improve them. We report them here:
\begin{itemize}
    \item \textbf{Winsorization:} given a CAM $V \in \mathbb{R}^{H \times W}$ and two values $l, h \in [0,100]$,
    we compute the corresponding percentiles $q_l$ and $q_h$. All pixel values below $q_l$ are raised to $q_l$,
    and all values above $q_h$ are lowered to $q_h$, ensuring a smoother normalization in the next step.
    \item \textbf{Normalization:} we then apply standard \textit{minmax} normalization to map the winsorized CAM into $[0,1]$.
    \item \textbf{Gamma correction:} to enhance contrast and better separate low-activation from high-activation regions,
    we apply a gamma correction $V^\gamma$ with coefficient $\gamma > 1$, followed by a renormalization.
\end{itemize}
Starting from the refined CAMs, we extract segmentation masks through \textbf{hysteresis thresholding}. In particular,
two percentiles are chosen as high and low thresholds: pixels above the high threshold are directly assigned to the mask, while pixels above the low threshold are included only if they are connected to already selected pixels.
With this procedure, we aim to produce coherent activation maps while reducing noise.

Bounding boxes are then extracted by computing the tightest axis-aligned rectangle that covers each connected component of the resulting masks.

It is worth remarking that the CAM refinement pipeline is robust across different activation distributions because it does not rely on any absolute threshold, but rather on quantiles.

This pipeline is used for YouTube-Objects v2.2, while for WS$^2$ we apply only minmax normalization followed by a fixed threshold $\tau=0.75$. The two pipelines differ because the benchmarks evaluate different outputs: CorLoc on YouTube-Objects v2.2 requires a bounding box, so hysteresis is used to obtain compact connected components from which boxes are extracted, whereas mIoU on WS$^2$ is computed directly on pixel masks, so a single global threshold on the normalized CAM suffices.

\section{Additional Ablation Studies}
\label{sec:ablation_supp}
As discussed in Section~\ref{sec:memcam}, our main contributions are the \textbf{MAM}, the \textbf{CAMPA}, and the \textbf{CAM Consistency Loss}.
In this section, we analyze the impact of these components on the overall performance of the model.
We also present results obtained under different training parameters and configurations, explaining how these experiments guided the selection of the final setup described in the previous subsection. Given the limited computational resources available, all ablation studies are conducted on a random subset of the original dataset. In particular, we first partition the videos into clips of length~$T$ according to the procedure described in Section~\ref{sec:supp_training}, then we select 30\% of the total clips uniformly at random. To prevent \texttt{OutOfMemory} errors during inference, we cap the resolution of test images to a maximum of $800\times800$ pixels while preserving the aspect ratio, and we run inference only at a scale of~0.7. Nonetheless, testing is still performed on the complete test set. Additionally, unless otherwise specified, all the MemCAM and MCTformer+ models used in this section feature a DeiT-Tiny backbone.

\subsection{Ablation of CAM Generation Hyperparameters}
\label{subsec:cam_ablation}

We ablate the hyperparameters of the \textbf{CAMPA} (Section~\ref{sec:cam_loss}).
The propagation is controlled by three parameters: the time decay \(\lambda\) (Eq. \ref{eq:time_decay}), the Top-\(K\) sparsification parameter \(K\) (Eq. \ref{eq:using_topK}), and the distribution temperature \(\mu\) (Eq. \ref{eq:using_temperature}).
Our declared objective is to produce \textbf{temporally consistent CAMs} while achieving good diffusion of CAM mass over object regions. Unless otherwise stated, we fix \(\lambda=0.97\).

\paragraph{Note on \(\lambda\).}
The temporal decay \(\lambda\) is not ablated systematically because its effect is strongly confounded by dataset-specific properties: an optimal value depends jointly on the video frame rate (which controls how much appearance changes between consecutive frames stored in memory) and the degree of scene staticness (slowly moving or nearly static objects tolerate a higher \(\lambda\), while fast-moving ones require faster decay). Both factors vary substantially across datasets and even across object classes within a dataset, making a controlled sweep uninformative without fixing these confounders. We therefore set \(\lambda\) empirically on a small validation subset and keep it fixed throughout all experiments.

\paragraph{Why \(K\) and \(\mu\) matter.}
For each current patch, the algorithm builds a probability distribution over patches from past frames using cosine similarity, and forms each synthetic CAM patch as a convex combination of past CAM patches weighted by this distribution.
\begin{itemize}
  \item \textbf{Low temperature (\(\mu\) small).} At low temperature, the distribution is peaked and fully concentrated around a few patches. Consequently, the convex combination becomes similar to nearest-neighbor matching, which in turn makes the effect of top-k sparsification negligible.
  \item \textbf{No sparsification (\(K=\infty\)).} The distribution covers all patches. If the temperature is high, the distribution approaches a uniform, so the convex combination turns into a global mean, coloring most of the image (see panel (b)).
  \item \textbf{High temperature + small \(K\) (best).}
  In this case, the distribution is flat but restricted only to the most similar patches. Consequently, similar current patches receive activations by similar sets of past patches, allowing uniformity across regions similar in feature space.
\end{itemize}

Balancing temporal consistency and spatial diffusion, we adopt
\[
\lambda=0.97,\qquad K=16,\qquad \mu=3.0
\]
for all training and testing.

\begin{figure}[t]
\centering
\includegraphics[width=\linewidth]{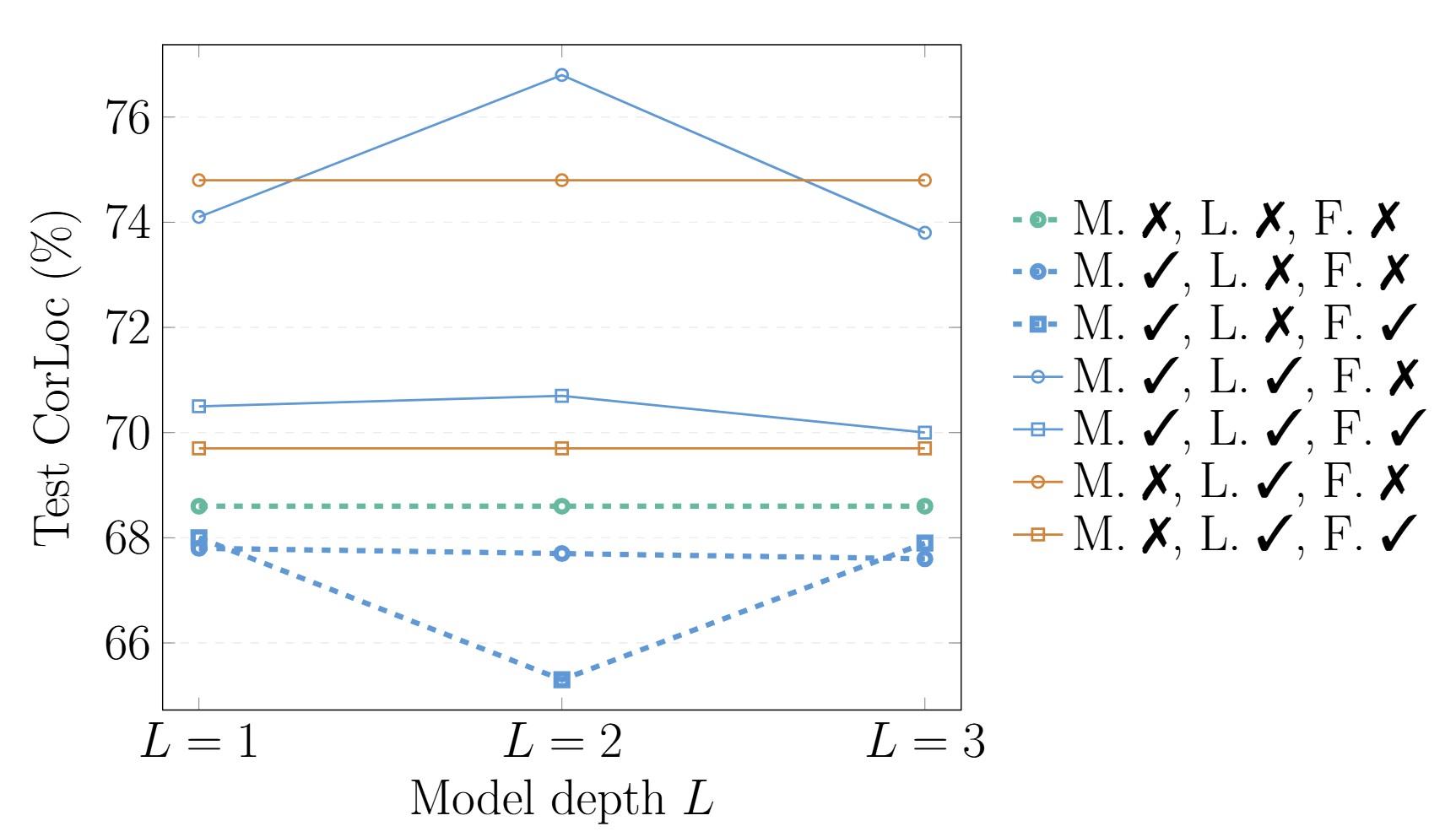}
\caption{Test CorLoc vs.\ model depth $L$.
\textbf{Encoding:} line style = loss configuration
(\textbf{solid}: $\lambda_{\mathrm{cls}}=0$, $\lambda_{p}=0$, $\lambda_{\mathrm{M}}=100$;
\textbf{dashed}: $\lambda_{\mathrm{cls}}=1$, $\lambda_{p}=1$, $\lambda_{\mathrm{M}}=0$);
color = memory (\textcolor{Blue}{\textbf{on}}, \textcolor{Red}{\textbf{off}}).
marker = backbone (\(\circ\) unfrozen, \(\square\) frozen).
The \textcolor{Green}{green} line indicates the baseline \textit{MCTformer+} model.
(M.=Memory, L.=CAM Loss, F.=Frozen)}
\label{fig:corloc_by_L}
\end{figure}

\subsection{Ablation of Model Configuration}
\label{subsec:memcam_config_ablation}

We ablate the architectural and training choices of \textbf{MemCAM}. The model is designed to be flexible: the \textbf{Memory Attention} block can contain an arbitrary number of Transformer decoder layers (depth \(L\)) or be disabled entirely, and training can use different combinations of Cross-Entropy and CAM Consistency losses via the weights \(\lambda_{\mathrm{cls}}, \lambda_{{p}}, \lambda_{\mathrm{M}}\) (\ref{eq:final_loss}).

\paragraph{Experimental setup.}
Unless otherwise noted, we initialize MemCAM from a baseline MCTformer+ trained from scratch: we load both backbone and head from the MCTformer+'s pretrained weights, and we leave the memory attention randomly initialized.
We compare two training regimes: freezing the MCTformer+ backbone and training only the Memory Attention and head and finetuning the entire network end-to-end.
We also compare different Memory Attention depths \(L\in\{1,2,3\}\) and two loss configurations: \emph{Consistency-only} (\(\lambda_{\mathrm{cls}}{=}0,\ \lambda_p{=}0,\ \lambda_{\mathrm{M}}{=}100\)) and \emph{CE-only} (\(\lambda_{\mathrm{cls}}{=}1,\ \lambda_{p}{=}1,\ \lambda_{\mathrm{M}}{=}0\)). The resulting test CorLoc is summarized in Fig.~\ref{fig:corloc_by_L}.

\begin{figure*}[t]
\centering

\begin{minipage}{0.19\textwidth}
    \centering
    \includegraphics[width=\linewidth]{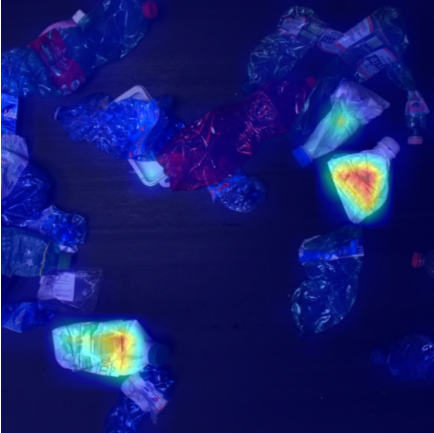}
    \caption*{GradCAM}
\end{minipage}\hspace{-0.3em}
\begin{minipage}{0.19\textwidth}
    \centering
    \includegraphics[width=\linewidth]{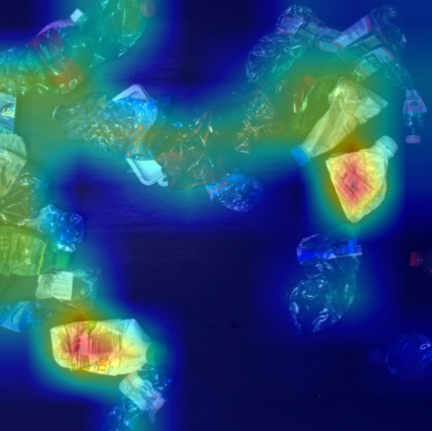}
    \caption*{GradCAM++}
\end{minipage}\hspace{-0.3em}
\begin{minipage}{0.19\textwidth}
    \centering
    \includegraphics[width=\linewidth]{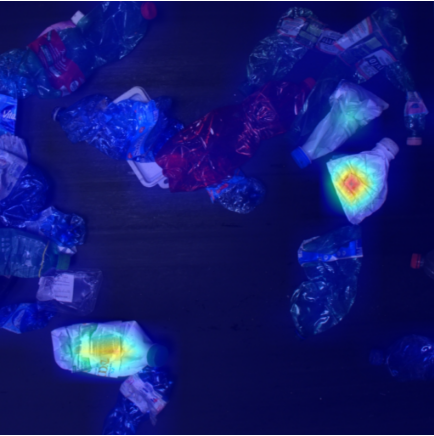}
    \caption*{LayerCAM}
\end{minipage}\hspace{-0.3em}
\begin{minipage}{0.19\textwidth}
    \centering
    \includegraphics[width=\linewidth]{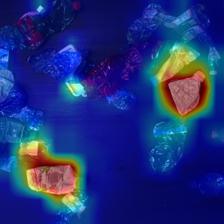}
    \caption*{MCTformer+}
\end{minipage}\hspace{-0.3em}
\begin{minipage}{0.19\textwidth}
    \centering
    \includegraphics[width=\linewidth]{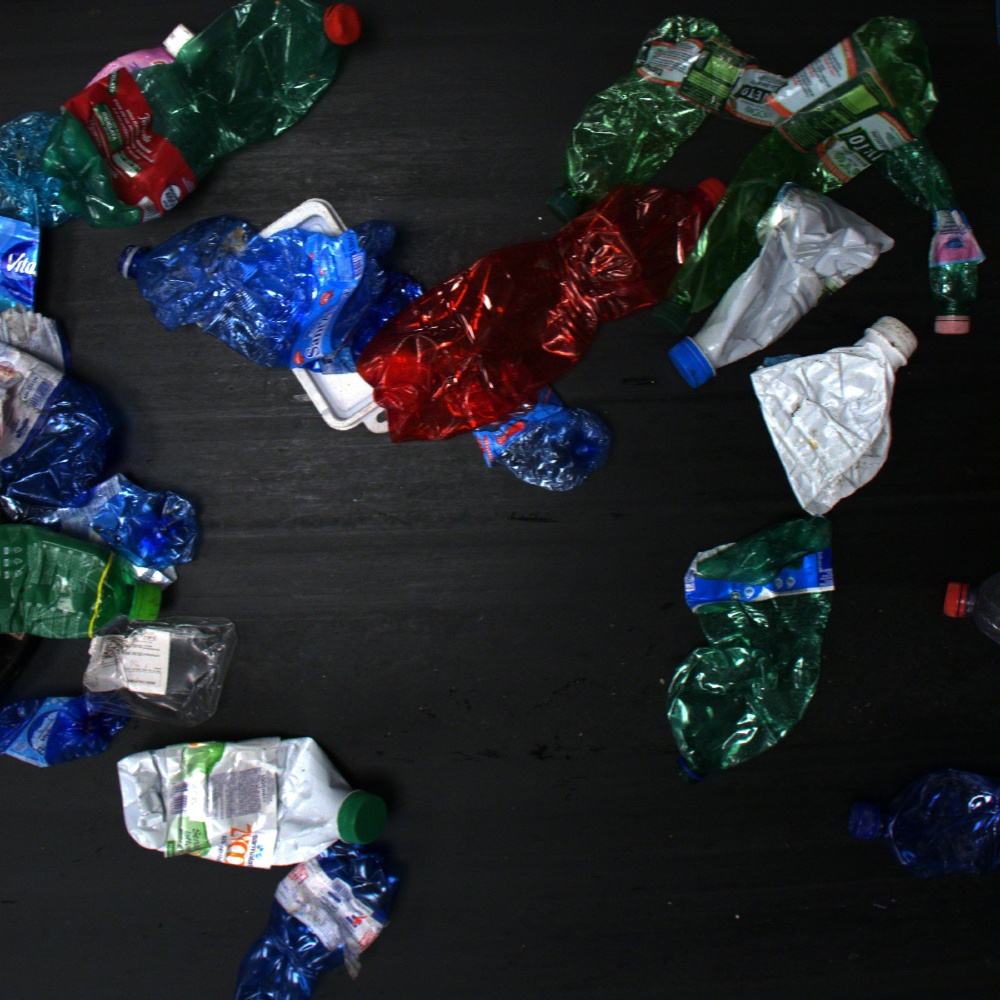}
    \caption*{Image}
\end{minipage}

\vspace{0.3cm}

\begin{minipage}{0.19\textwidth}
    \centering
    \includegraphics[width=\linewidth]{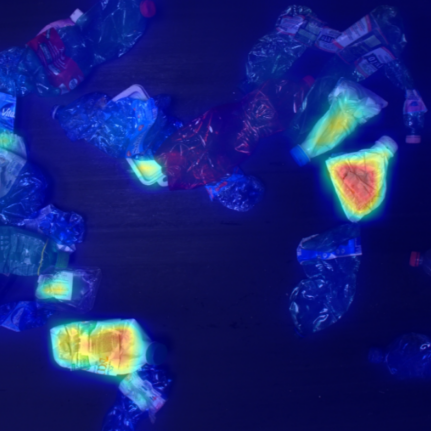}
    \caption*{PuzzleCAM}
\end{minipage}\hspace{-0.3em}
\begin{minipage}{0.19\textwidth}
    \centering
    \includegraphics[width=\linewidth]{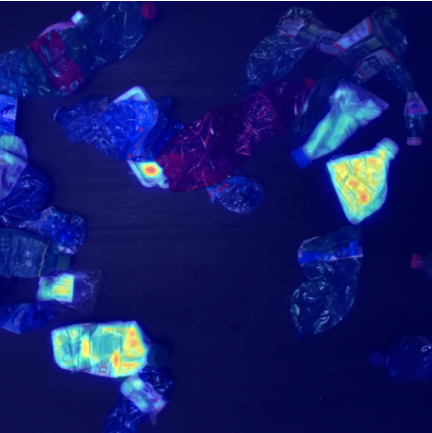}
    \caption*{POF-CAM}
\end{minipage}\hspace{-0.3em}
\begin{minipage}{0.19\textwidth}
    \centering
    \includegraphics[width=\linewidth]{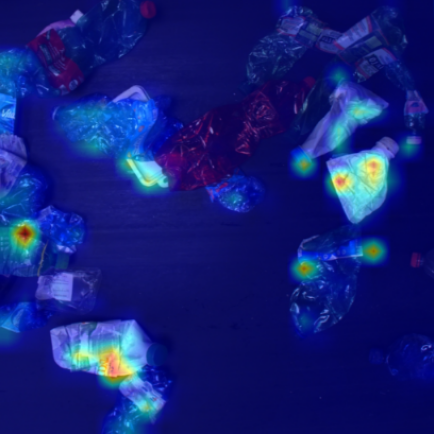}
    \caption*{WeakTr}
\end{minipage}\hspace{-0.3em}
\begin{minipage}{0.19\textwidth}
    \centering
    \includegraphics[width=\linewidth]{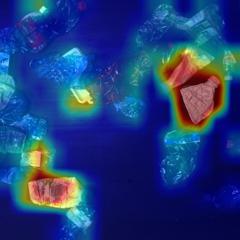}
    \caption*{MemCAM}
\end{minipage}\hspace{-0.3em}
\begin{minipage}{0.19\textwidth}
    \centering
    \includegraphics[width=\linewidth]{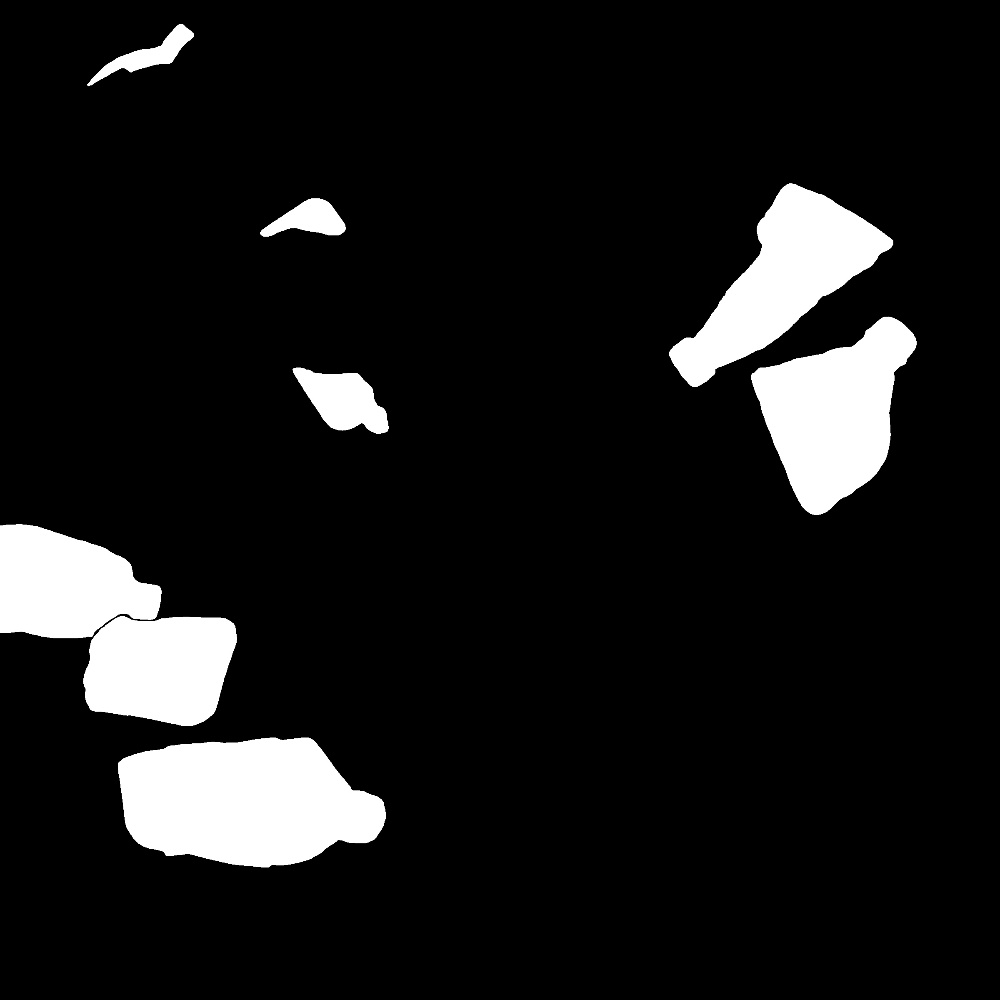}
    \caption*{Ground Truth}
\end{minipage}

\caption{
Qualitative comparison of CAMs produced by different weakly supervised localization methods on a WS$^2$ test sample.
}
\label{fig:cams_of_different_types}
\end{figure*}

\paragraph{Results.}
From Fig.~\ref{fig:corloc_by_L} we draw four conclusions:
\begin{enumerate}

    \item The addition of Memory Attention and Memory Bank without an auxiliary loss does \emph{not} improve the model's performance with respect to the baseline. We deduce that without an explicit signal, the model does not learn to effectively use its memory for localization.

  \item The introduction of the \textit{CAM Consistency Loss} while disabling other losses provides a significant CorLoc gain (\(+1.5\%\) over the baseline). We deduce that this loss alone is responsible for an improvement in localization capabilities.

  \item When the Consistency Loss is active, enabling the Memory Attention yields additional improvements, from \(\sim\!+0.5\%\) when the backbone is frozen to nearly \(+2\%\) when it is unfrozen, signaling that Memory and CAM Loss work in synergy to improve the model's performance.

  \item Training with a frozen backbone limits adaptability and leads to suboptimal results.
\end{enumerate}

\paragraph{Retained configuration.}
Balancing accuracy and stability, we adopt the end-to-end setting with an unfrozen backbone, the Memory Attention depth \(L{=}2\), and the \emph{Consistency-only} loss weights
\[
\lambda_{\mathrm{cls}}=0,\qquad \lambda_{p}=0,\qquad \lambda_{M}=100.
\]
This configuration attains a test CorLoc of \(76.8\%\) and a test mIoU of \(59.2\)\%, outperforming all other variants across all splits.

\subsection{Ablation of Data Augmentations}
\label{subsec:aug_ablation}

We now study the influence of standard image augmentations on MemCAM\_tiny. All experiments adopt the retained training configuration described in Section~\ref{subsec:memcam_config_ablation}. We evaluate the effect of enabling or disabling two commonly used transformations, \textit{ColorJitter} and \textit{Random Erasure}, and test all four possible combinations. The goal of this analysis is to understand how appearance perturbations (ColorJitter) and local occlusions (Erasure) interact with temporal consistency and CAM propagation in a weakly supervised setting. The resulting performance variations are summarized in Table~\ref{tab:augs_results30}.

\paragraph{Experimental setup.}
Each model is trained on 30\% of the dataset using the same optimizer, learning-rate schedule, and loss configuration as in Section~\ref{subsec:memcam_config_ablation}. The only change across runs is the activation of the two augmentations: no augmentation, ColorJitter only, Erasure only, or both together. This controlled setup ensures that differences in performance can be attributed solely to the augmentations.

\begin{table}[t]
\centering
\begin{adjustbox}{width=\columnwidth}
\begin{tabular}{cccc}
\toprule
\textbf{ColorJitter} & \textbf{Erasure} & \textbf{Test CorLoc (\%)} & \textbf{Test mIoU (\%)}\\
\midrule
\xmark & \xmark & 73.6 & 58.0 \\
\cmark & \xmark & 75.5 & 59.0 \\
\xmark & \cmark & 72.4 & 56.6 \\
\cmark & \cmark & \textbf{76.8} & \textbf{59.2} \\
\bottomrule
\end{tabular}
\end{adjustbox}
\caption{Effect of ColorJitter and Random Erasure on MemCAM\_tiny trained on 30\% of the data. Best scores in bold.}
\label{tab:augs_results30}
\end{table}

\paragraph{Results.}
Several trends emerge from this ablation. First, \textbf{ColorJitter alone provides consistent benefits}, raising CorLoc by \(+1.9\) points (73.6\(\rightarrow\)75.5) and mIoU by \(+1.0\) point (58.0\(\rightarrow\)59.0), suggesting that mild appearance variability strengthens the model’s spatial generalization. Conversely, \textbf{Erasure alone degrades performance}, reducing CorLoc and mIoU by \(-1.2\) and \(-1.4\) points, respectively, likely because patch-level masking conflicts with the CAM propagation signal. Interestingly, combining both transformations yields the \textbf{best overall performance} (76.8\% CorLoc, 59.2 mIoU), indicating that the regularizing effect of ColorJitter compensates for the aggressiveness of Erasure, and that the two augmentations jointly act as a more effective training signal than either in isolation.

\subsection{Classification Accuracy and Domain Knowledge}
\label{subsec:supp_classification}

CAM-based methods are commonly assumed to inherit the quality of the classifier they are extracted from, so that better recognition should imply better localization. This is in fact a known problem in weakly supervised localization~\cite{choe2020evaluating, kim2022bridging, guichemerre2024source}: the classification objective is minimized by the most \emph{discriminative} evidence rather than by the full object extent, so recognition accuracy is neither a reliable proxy for localization quality nor a sound criterion for model selection. Our setting allows the assumption to be tested directly, because the retained configuration trains the student \emph{without} any classification objective. We therefore measured top-1 accuracy for both the frozen teacher and the trained student on the two benchmarks, and set it against the corresponding localization and segmentation scores.

The measurements are reported in Table~\ref{tab:classification_vs_localization}. Unlike the other ablations of this section, they are obtained under the full evaluation protocol of the main paper, so that the localization scores match those of Tab.~2 and Tab.~3. The teacher reaches $94.7\%$ classification accuracy on WS$^2$ and the student $85.9\%$; on YTOv2.2 the teacher reaches $93.2\%$ and the student $87.6\%$. Localization improves on both datasets ($+4.3$ mIoU and $+8.2$ CorLoc) while classification \emph{decreases} on both ($-8.8$ and $-5.6$ points), showing that in our setting higher classification accuracy is not necessary for better CAM localization, coherently with the network loss design exposed in Sec.~\ref{sec:ablation} and Sec.~\ref{subsec:memcam_config_ablation}. The mechanism follows directly from the retained configuration. With $\lambda_{\mathrm{cls}}{=}\lambda_{p}{=}0$ (Sec.~\ref{subsec:memcam_config_ablation}) the student receives no classification gradient at all: the class-token branch is inherited from the teacher initialization and is then free to drift, while the consistency loss reshapes the spatial activations. What matters for CAM quality in our architecture is therefore not how accurately the network names the scene, but where it looks when it does.

\paragraph{How much domain knowledge does the backbone need?}
The two benchmarks differ sharply in how much of their content is covered by the backbone's pre-training. YTOv2.2 contains ten common object categories that ImageNet represents well, whereas WS$^2$ consists of industrial recycling scenes, with irregular waste objects on a conveyor belt, supervised only by \emph{before}/\emph{after} scene labels; it is a domain essentially absent from ImageNet, and it is precisely the regime in which methods that inherit frozen semantics from a foundation model degrade. Since MemCAM yields comparable gains over its teacher both on a dataset whose categories are covered by ImageNet pre-training (YTOv2.2) and on one where they are not (WS$^2$), our results suggest that exact domain knowledge at pre-training time is not necessary, provided a video-level classifier can be fitted in-domain. We note, however, that the backbone remains ImageNet-pretrained, so these experiments cannot establish whether general-purpose pre-training knowledge is itself dispensable.

The requirement CAMPA places on the backbone is therefore weaker than it may appear, and it is a requirement of \emph{trainability} rather than of prior coverage: what the method needs is a feature space in which semantically related patches of nearby frames lie close, and a per-class activation map to propagate (Sec.~\ref{sec:supp_scope}). Both are obtained by fitting the weak classifier in-domain, which is the minimal assumption of the weakly supervised setting itself. Propagation quality is then upper-bounded by the resulting CAMs, which is the failure mode analysed in Sec.~\ref{sec:supp_failure}.

\begin{table}[t]
\centering
\begin{adjustbox}{width=\columnwidth}
\begin{tabular}{lcccc}
\toprule
& \multicolumn{2}{c}{\textbf{YTOv2.2}} & \multicolumn{2}{c}{\textbf{WS$^2$}} \\
\cmidrule(lr){2-3}\cmidrule(lr){4-5}
\textbf{Model} & \textbf{Acc. (\%)} & \textbf{CorLoc (\%)} & \textbf{Acc. (\%)} & \textbf{mIoU (\%)} \\
\midrule
MCTformer+ (teacher) & 93.2 & 68.6 & 94.7 & 36.90 \\
MemCAM (student)     & 87.6 & \textbf{76.8} & 85.9 & \textbf{41.21} \\
\midrule
$\Delta$ & $-5.6$ & $\mathbf{+8.2}$ & $-8.8$ & $\mathbf{+4.3}$ \\
\bottomrule
\end{tabular}
\end{adjustbox}
\caption{Classification accuracy against localization and segmentation quality, for the frozen teacher and the MemCAM student on both benchmarks, with the DeiT-Tiny backbone. Large localization gains are obtained while classification accuracy decreases on both benchmarks.}
\label{tab:classification_vs_localization}
\end{table}

\subsection{Ablation of the Memory Length \texorpdfstring{$T$}{T}}
\label{subsec:supp_memory_length}

The memory length $T$ controls how far back CAMPA can look: it sets the number of past frames stored in both queues, the number of similarity blocks in $S$, and therefore the storage cost, which grows linearly with $T$. We sweep $T\in\{2,3,5,7,10\}$ under the protocol of this section, retraining the model for each value.

Since a subclip consists of the $T$ frames that populate the memory plus the current one, changing $T$ changes the clip length to $T+1$. To keep the comparison controlled, for each $T$ we choose the window length $L_w$ as the multiple of $T+1$ closest to $30$, so that every window spans approximately the same number of frames and only the memory horizon varies.

\begin{table}[t]
\centering
\begin{adjustbox}{width=\columnwidth}
\begin{tabular}{ccccc}
\toprule
$\mathbf{T}$ & \textbf{Clip len.} & \textbf{$L_w$} & \textbf{Test CorLoc (\%)} & \textbf{Test mIoU (\%)} \\
\midrule
2  & 3  & 30 & 75.7 & 59.0 \\
3  & 4  & 32 & 76.2 & \textbf{59.4} \\
5  & 6  & 30 & \textbf{76.4} & 59.3 \\
7  & 8  & 32 & 76.0 & 58.6 \\
10 & 11 & 33 & 74.2 & 57.5 \\
\bottomrule
\end{tabular}
\end{adjustbox}
\caption{Effect of the memory length $T$ on YTOv2.2, with the DeiT-Tiny backbone. Performance is flat over $T\in\{2,3,5,7\}$ and degrades at $T{=}10$. Best scores in bold.}
\label{tab:memory_length}
\end{table}

\paragraph{Results.}
Table~\ref{tab:memory_length} shows that the method is \emph{not} sensitive to $T$ over the useful range: between $T{=}2$ and $T{=}7$ CorLoc varies by $0.7$ points and mIoU by $0.8$, a spread comparable to run-to-run noise. Performance degrades only at $T{=}10$ ($-2.2$ CorLoc, $-1.9$ mIoU with respect to the best setting), which is consistent with the failure mode analysed in Sec.~\ref{sec:supp_failure}: as the window widens, the oldest frames in memory are increasingly distant in appearance from the current one, their cross-frame affinities become unreliable, and the temporal decay $\lambda^{\tau}$ down-weights them but does not remove them from the Top-$K$ candidate pool. A longer memory therefore adds candidates without adding usable evidence.

The flatness over the useful range is itself informative: it indicates that the gain of CAMPA does not come from aggregating many frames, but from aggregating a few \emph{reliable} correspondences, which is also why the Top-$K$ sparsification of Sec.~\ref{subsec:cam_ablation} matters more than the window size. We retain $T=5$, which attains the best CorLoc while keeping the Memory Bank small.

\section{Additional Analysis on the WS$^2$ Dataset}
\label{sec:supp_ws2}

This section expands the discussion introduced in the main paper's Experiments section (Sec.~\ref{sec:datasets}), providing a more detailed analysis of the WS$^2$~\cite{marelli2025ws2} dataset and of MemCAM's behaviour in this challenging setting. WS$^2$ is the only existing benchmark specifically designed for Weakly Supervised Video Semantic Segmentation and consists of approximately \textbf{11k frames} captured in an industrial recycling plant, divided into two scene-level classes: \textit{before} and \textit{after}. The \textit{before} frames depict a conveyor belt containing both \emph{wanted} and \emph{unwanted} objects, whereas the \textit{after} frames contain only the \emph{wanted} ones. The task is formulated as binary scene classification, where CAMs extracted from the classifier are used to segment the \emph{unwanted objects} (recyclables) that appear exclusively in the \textit{before} frames. This implies that the model must learn, without pixel-level supervision, that “\textit{before} = presence of unwanted objects’’ and “\textit{after} = absence of unwanted objects,’’ making WS$^2$ an inherently difficult and highly domain-specific benchmark.

\subsection{Dataset properties and intrinsic difficulty}

WS$^2$ is small and highly repetitive: videos are recorded from fixed viewpoints, under controlled lighting, and with nearly identical backgrounds.  Moreover, \textit{after} frames do not contain any positive visual evidence, only the \emph{absence} of recyclable material differentiates them from \textit{before} frames.
This makes WS$^2$ much harder than general WS Segmentation image datasets: CAM-based models must learn to activate only where unwanted objects appear, learning this only from their presence in \emph{before} images \cite{marelli2025ws2}.  As in the original WS$^2$ protocol, we train MemCAM and MCTformer+ on images where the background region is masked out, ensuring a fair comparison with prior work.

\subsection{Quantitative analysis}
As reported in Table~\ref{tab:miou_table_compact}, MemCAM\_tiny achieves the best mIoU (41.21\%), outperforming all image-level WS methods (GradCAM \cite{selvaraju2017grad}, GradCAM++ \cite{chattopadhay2018grad}, LayerCAM \cite{jiang2021layercam}, PuzzleCAM \cite{jo2021puzzle}) as well as temporal baselines relying on external cues such as optical flow (POF-CAM \cite{marelli2024temporal}).
Importantly, it also surpasses its own backbone, MCTformer+\_tiny (36.90\%), demonstrating the benefit of temporal refinement in the latent space.

The larger MemCAM\_small struggles (25.03\%), similarly to MCTformer+\_small (31.13\%).
This confirms that the dataset is too small and repetitive for high-capacity transformers, which tend to overfit background textures and fail to consistently highlight the small, irregular recyclable objects.
In this setting, compact architectures, such as MemCAM\_tiny, strike the right balance between capacity and generalization, enabling stable temporal refinement and leading to state-of-the-art performance.

\subsection{Summary}

Despite the limited size and strong domain bias of WS$^2$, MemCAM\_tiny establishes a new state of the art.
The results highlight two main points:

\begin{itemize}
    \item \textbf{Appropriate model scaling is critical}: larger Transformers overfit and degrade, while MemCAM\_tiny effectively exploits temporal consistency to refine CAMs under scarce supervision.
    \item \textbf{Feature-level temporal reasoning is effective even without external motion cues}: MemCAM learns cross-frame correspondences directly in the latent space, outperforming optical-flow-based approaches and all image-only baselines.
\end{itemize}

Overall, WS$^2$ demonstrates that MemCAM is robust to domain-specific constraints and highly competitive in real industrial segmentation scenarios.

\section{Disentangling Temporal Propagation from Self-Distillation}
\label{sec:supp_distill}

This section expands the discussion deferred from the main paper (Sec.~\ref{sec:datasets}). Because MemCAM is trained with a frozen teacher (Sec.~\ref{sec:supp_training}), one may ask whether the gain stems from generic teacher--student regularization rather than from temporal propagation. Three observations rule this out. First, the teacher is simply the frozen single-frame baseline, and the student is never trained to match the teacher's prediction \emph{on the same frame}; the only signal tying them is the CAM consistency loss, whose target is the CAMPA output. Second, that target is not the teacher's frame-$t$ CAM but a cross-frame, similarity-weighted aggregation of \emph{past} CAMs (Eq.~\ref{eq:propagated_cam}): a pure self-distillation signal is upper-bounded by the teacher and cannot, by construction, push the student above it, yet MemCAM improves over the teacher by $+8.2$ CorLoc ($68.6\!\rightarrow\!76.8$, Table~\ref{tab:loss_ablation}). Third, the ablation isolates the two effects: with the propagated supervision but \emph{without} the MAM ($74.8$, i.e.\ $+6.2$) the gain is carried by the temporal target alone, and adding the learnable cross-attention contributes a further $+2.2$. Both the temporal target and the temporal weights, not distillation, therefore drive the result.

\subsection{Using the propagated CAM as the final prediction}
\label{subsec:supp_campa_as_pred}

A natural question follows from the design: if $\hat{M}^t$ is a good enough target to supervise the network, why not use it directly as the model output, dispensing with $M_c^t$ altogether? We also evaluated the propagated CAM $\hat{M}^t$ as the final prediction. The map of Eq.~\ref{eq:propagated_cam} is fed to the \emph{same} post-processing pipeline used for $M_c^t$ (Sec.~\ref{sec:supp_training}), so that the two variants differ only in which map enters it. Both predictors are scored on identical data, namely the frames for which the memory is already populated, since $\hat{M}^t$ is undefined on the first $T$ frames of each video. We report the \emph{autoregressive} setting, in which the student fills its own memory with the CAMs it has produced: this is the realistic deployment condition, as opposed to training, where the memory is filled by the frozen teacher (Alg.~\ref{alg:training_protocol}). All models use the DeiT-Tiny backbone.

\begin{table}[t]
\centering
\begin{adjustbox}{width=\columnwidth}
\begin{tabular}{lcc}
\toprule
\textbf{Map used as prediction} & \textbf{YTOv2.2 CorLoc (\%)} & \textbf{WS$^2$ mIoU (\%)} \\
\midrule
Source CAMs (frozen teacher)            & 68.6          & 36.90 \\
MemCAM prediction $M_c^t$ \emph{(ours)} & 76.8          & 41.21 \\
Propagated $\hat{M}^t$ (CAMPA output)   & \textbf{77.5} & \textbf{41.62} \\
\bottomrule
\end{tabular}
\end{adjustbox}
\caption{The propagated map $\hat{M}^t$ used directly as the final prediction, against the CAMs it aggregates and against the student prediction. Propagation improves substantially over its own inputs and then matches the learned predictor. Both predictors are scored on the frames whose memory is populated.}
\label{tab:campa_as_prediction}
\end{table}

The outcome, reported in Table~\ref{tab:campa_as_prediction}, follows a clear two-step pattern. It improves over the source CAMs it aggregates by $+8.9$ CorLoc on YTOv2.2 and $+4.7$ mIoU on WS$^2$, confirming its value as a supervisory target, and it matches the student prediction to within $0.7$ CorLoc and $0.4$ mIoU respectively. Propagation thus removes a large part of the per-frame noise of the source maps, and the residual margin between $\hat{M}^t$ and the student is far below the gap that separates either of them from those maps.

We regard this agreement as the intended outcome rather than as a reason to replace the prediction: the consistency loss $\mathcal{L}_{\mathrm{M}}$ explicitly minimises the divergence between $M_c^t$ and $\hat{M}^t$, so at convergence the two maps \emph{must} coincide, and their equivalence is direct evidence that the temporal supervision has been internalised by the network rather than that the student is redundant. This is consistent with the role of $\hat{M}^t$ as a supervisory target: CAMPA exploits temporal redundancy during training, while the student internalizes this supervision and recovers the same accuracy through the standard inference path, without requiring past CAMs or feature-matching at inference.

The choice between the two is therefore one of cost, not of quality. By Eq.~\ref{eq:propagated_cam}, $\hat{M}^t$ is a function of the past CAMs the model itself produces, so using it as the output leaves the first $T$ frames of every video without a prediction, and it requires retaining the CAM queue and recomputing the $N^2\!\times\!(T N^2)$ similarity at every frame, whereas the student maps one frame to a mask in a single forward pass. CAMPA makes the temporal redundancy of the data explicit, and training distils it into the weights, so that the network reproduces it from a single frame without the memory that produced it. This is also why the student is not bounded by what the aggregation can express: a convex combination of past CAMs cannot introduce evidence that is absent from memory, such as an object entering the scene at time $t$, yet the student exceeds the frozen teacher that generates those targets by $+8.2$ CorLoc.

\paragraph{Learned temporal correspondences.}
Beyond final metrics, the qualitative sequences in Fig.~\ref{fig:qualitative_comparison_seq} show that MemCAM's activations are markedly more stable across consecutive frames than the static MCTformer+ baseline, with fewer flickering background activations and more consistent object coverage, indicating that the learned cross-attention captures coherent inter-frame associations rather than per-frame noise. We note that a direct, quantitative evaluation of correspondence accuracy would require dense point-tracking ground truth, which neither weakly labeled benchmark provides; we therefore report correspondence quality indirectly, through temporal-consistency visualizations and the downstream localization and segmentation gains. The failure modes of this matching are reported in Sec.~\ref{sec:supp_failure}.

\section{Backbone Requirements and Scope of MemCAM}
\label{sec:supp_scope}

This section details the requirements that CAMPA imposes on the backbone and clarifies which architectures it applies to, expanding the scope discussion deferred from the main paper (Sec.~\ref{sec:datasets}).

CAMPA assigns the backbone two distinct roles. First, it must produce \emph{patch tokens} whose pairwise affinities yield reliable cross-frame correspondences: this is what allows the similarity matrix $S$ (Eq.~\ref{eq:comp_sim_submatrixes}) to identify, for each current patch, the semantically related patches in past frames. Second, it must expose \emph{per-class CAMs} that can be stored, propagated, and supervised: these are the quantity actually transferred across time in Eq.~\ref{eq:propagated_cam}. The two roles are independent: the first only needs a feature space in which semantically related patches lie close, whereas the second additionally needs a class-discriminative activation map to transfer.

The MCTformer+ family~\cite{xu2024mctformer+, xu2022multi, gao2021ts, zhu2023weaktr} fulfils both roles natively through its multi-class-token design: the patch tokens provide the embedding space used for matching, while each class token yields a class-discriminative activation map. This is precisely why we adopt it. We stress that the constraint concerns \emph{what} CAMPA propagates, namely the need for a per-class CAM to transfer across frames, and not the quality of the feature matching, which any sufficiently strong patch encoder already provides.

This requirement also clarifies why generic self-supervised encoders are not drop-in replacements. Models such as DINO~\cite{caron2021emerging, simeoni2025dinov3} excel at the first role, providing strong patch correspondences, but expose no class-discriminative CAM head and therefore cannot satisfy the second: there is no per-class activation to propagate or supervise. Integrating such features would require attaching and weakly training a dedicated token-CAM head on top of the (possibly frozen) representations and re-deriving CAMPA's targets from it, effectively \emph{decoupling} the correspondence space from the CAM space. We regard this decoupling as a promising research direction in its own right (left as future work in the conclusion of the main paper) rather than a hyperparameter swap, and we therefore deliberately keep the correspondence and CAM spaces tied to a single backbone in the present work, so as to isolate the effect of the proposed temporal supervision.

\section{Failure Cases and Limitations}
\label{sec:supp_failure}

This section expands the failure-case analysis deferred from the main paper (Sec.~\ref{sec:datasets}). CAMPA inherits two limitations from its feature-matching core.

\paragraph{Propagation is bounded by the teacher.}
Because the synthetic targets are synthesized from the teacher's past CAMs (Eq.~\ref{eq:propagated_cam}), propagation can only be as good as those activations. On highly repetitive, low-diversity data the teacher itself is weak, so propagating weak CAMs yields limited gains. This is exactly the regime we observe on the high-capacity WS$^2$ variant (Sec.~\ref{sec:supp_ws2}), where an over-parameterized MCTformer+ teacher overfits the near-static background and its degraded CAMs cap the quality of the propagated targets. Aligning backbone capacity with dataset scale, as in MemCAM\_tiny, is the practical mitigation, consistent with the capacity study of the main paper (Sec.~\ref{sec:datasets}).

\paragraph{Affinities degrade under abrupt appearance change.}
When appearance changes abruptly within the memory window, e.g.\ fast motion, strong occlusion, or shot-cut-like transitions, the cross-frame affinities of Eq.~\ref{eq:comp_sim_submatrixes} become unreliable and may match semantically unrelated patches. Two design choices mitigate this: the Top-$K$ filtering and temporal decay of CAMPA (Sec.~\ref{sec:supp_campa}) suppress spurious and distant-in-time matches, while the dual-residual design of the MAM (Eq.~\ref{eq:supp_final_residual}) lets the model fall back on the current-frame features when the retrieved memory is uninformative. These mechanisms reduce, but do not fully eliminate, the sensitivity to large inter-frame displacements; handling such cases explicitly, e.g.\ with motion-aware memory selection, is left to future work.

\paragraph{The memory horizon is short by construction.}
The Memory Bank is a FIFO queue of the $T$ most recent frames, deliberately the simplest possible policy, so that the observed gain is attributable to the propagated supervision rather than to a memory-selection heuristic. The horizon is therefore bounded by $T$: objects that leave the scene and return, and dependencies spanning more than $T$ frames, are not captured.

\paragraph{Why other video benchmarks cannot be adapted.}
Our protocol requires a dataset that provides (i) a \emph{semantic class taxonomy}, so that a single video-level class label can be assigned to each clip; (ii) enough videos to train a classifier from which CAMs can be extracted; and (iii) per-frame localization or segmentation ground truth for \emph{evaluation only}, ideally accompanied by existing weakly supervised baselines for comparison. Only two public benchmarks satisfy all three, YTOv2.2 (general domain) and WS$^2$ (industrial), and established video segmentation benchmarks do not. Object-centric video segmentation datasets such as DAVIS annotate \emph{class-agnostic} foreground or instance masks, with no semantic taxonomy from which a video-level class label could be derived, and contain only a few dozen videos, far too few to train a CAM classifier; they are moreover built for semi-supervised mask propagation, where the first-frame mask is given, a different task for which no weakly supervised \emph{semantic} protocol or baseline exists. The Actor-Action Dataset (A2D) is the closest candidate: it provides a semantic taxonomy of actor classes and sparse per-frame pixel masks, but a substantial portion of its videos contains several actor classes at once, which breaks the single-label formulation of Sec.~\ref{sec:problem}. Its single-label subset would satisfy (i)--(iii), whereas the full multi-label setting would require propagating one CAM channel per class present in the clip; we regard both as a natural extension and leave them to future work. Action-recognition datasets such as Something-Something carry clip-level \emph{action} labels that do not correspond to segmentable object classes. Adapting any of these would mean defining a new task and evaluation protocol from scratch, with no prior numbers to compare against, which is a benchmarking contribution orthogonal to our method. Building such a larger, general-purpose WS-VSS benchmark is an important community need that we leave to future work. Consistently with each benchmark's established protocol, we report CorLoc on YTOv2.2 (localization, as no per-pixel ground truth exists) and mIoU on WS$^2$ (segmentation).